\PassOptionsToPackage{round,authoryear}{natbib}

\documentclass[preprint,12pt]{elsarticle}

\usepackage{amssymb}
\usepackage{amsmath}
\usepackage{threeparttable}
\usepackage{multirow}
\usepackage{booktabs}
\usepackage{float} 
\usepackage{arydshln}
\usepackage[left=1.5cm, right=1.5cm, top=2cm, bottom=2cm]{geometry}
\usepackage{setspace}
\usepackage[round]{natbib}
\usepackage[colorlinks=true, linkcolor=blue, citecolor=blue, urlcolor=blue]{hyperref}

\begin{document}

\begin{frontmatter}

%\title{Comprehensive Analysis of Power Consumption Across ICE, EV, and HEV Using Deep Learning}
\title{Deep Learning-Based Analysis of Power Consumption in Gasoline, Electric, and Hybrid Vehicles}

\author[1]{Roksana Yahyaabadi} %% Author name
\author[1]{Ghazal Farhani} 
\author[1]{Taufiq Rahman} 
\author[2]{Soodeh Nikan} 
\author[1]{Abdullah Jirjees} 
\author[3]{Fadi Araji} 
\affiliation[1]{organization={National Research Council Canada},%Department and Organization
            city={London},
            state={Ontario},
            country={Canada}}
 \affiliation[2]{organization={Department of Electrical and Computer Engineering, Western University},%Department and Organization
            city={London},
            state={Ontario},
            country={Canada}}
 \affiliation[3]{organization={Environmnet Climate Change Canada},%Department and Organization
            city={Ottawa},
            state={Ontario},
            country={Canada}}

%% Abstract
\begin{abstract}

Accurate power consumption prediction is crucial for improving efficiency and reducing environmental impact, yet traditional methods relying on specialized instruments or rigid physical models are impractical for large-scale, real-world deployment. This study introduces a scalable data-driven method using powertrain dynamic feature sets and both traditional machine learning and deep neural networks to estimate instantaneous and cumulative power consumption in internal combustion engine (ICE), electric vehicle (EV), and hybrid electric vehicle (HEV) platforms. ICE models achieved high instantaneous accuracy with mean absolute error and root mean squared error in order of $10^{-3}$, and cumulative errors under 3\%. Transformer and long short-term memory performed best for EVs and HEVs, with cumulative errors below 4.1\% and 2.1\%, respectively. Results confirm the approach’s effectiveness across vehicles and models. Uncertainty analysis revealed greater variability in EV and HEV datasets than ICE, due to complex power management, emphasizing the need for robust models for advanced powertrains.
\end{abstract}


% \begin{highlights}
% \end{highlights}

%% Keywords

\end{frontmatter}

\section{Introduction}\label{Introduction}

The transportation sector is one of the largest global energy consumer and a significant contributor to air pollution. It accounts for more than 30\% of total delivered energy, with road vehicles alone responsible for almost 80\% of the sector's energy consumption \citep{ullah2022comparative}. Given this substantial energy demand, improving fuel and energy efficiency is critical in all types of vehicles. Improving fuel efficiency in internal combustion engine (ICE) vehicles is crucial to reducing emissions and operational costs. Meanwhile, electrical vehicles (EVs) and hybrid electric vehicles (HEVs) benefit from optimized power consumption to enhance performance and sustainability. These vehicles employ strategies such as regenerative braking and power optimization to further improve efficiency and reduce reliance on fossil fuels, contributing to the wider transition towards sustainable transportation. Effective energy management is essential for all types of vehicles. In this context, accurate modeling of the vehicle's power consumption enables the identification of driving strategies that optimize energy efficiency, reduce costs, and minimize emissions. Furthermore, real-time power consumption monitoring not only improves fuel efficiency but also aligns with global initiatives to achieve net-zero emissions by 2050 and meet sustainable development goal 13 \citep{zhao2023review}.

\begin{figure*}[!t]
\centering
\includegraphics[width=\linewidth]{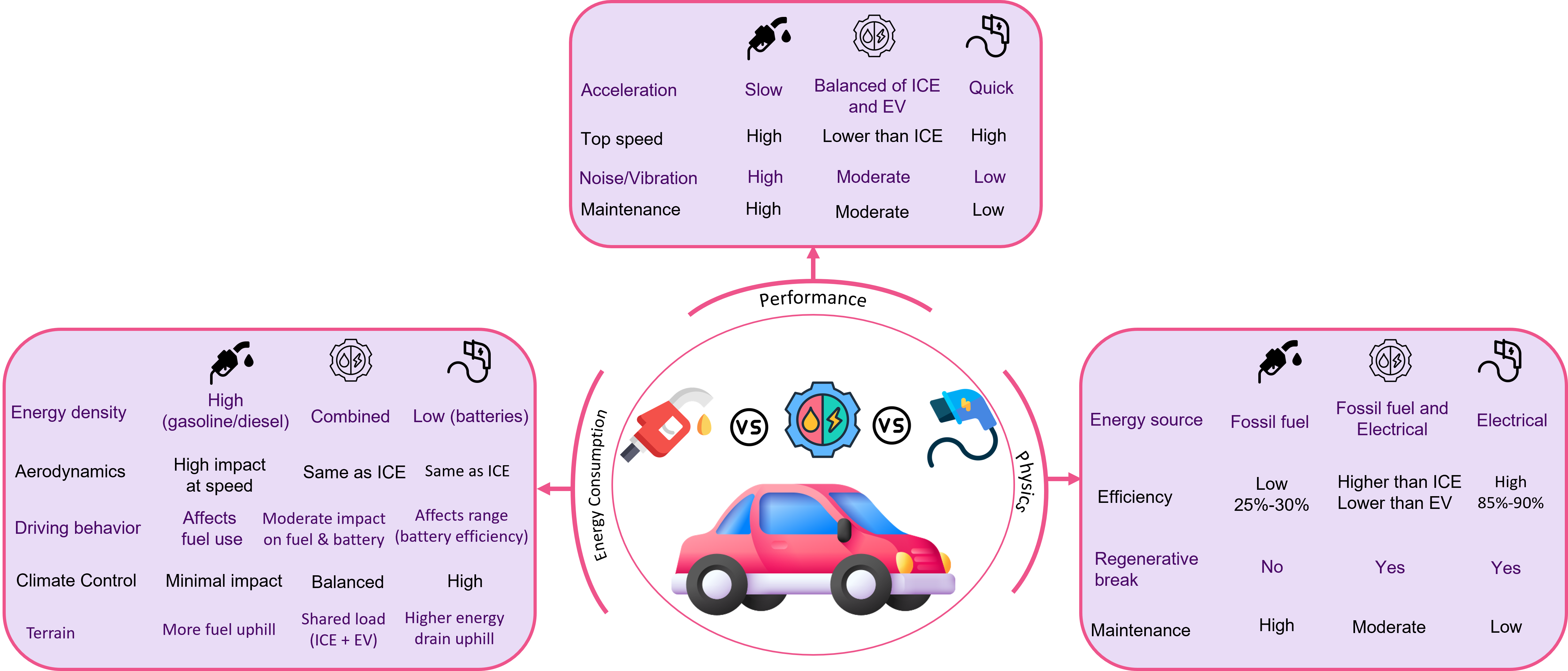}
\caption{Comparative overview of the main characteristics of ICE vehicles, EVs and HEVs in terms of physics, performance and power consumption.}
\label{ComparisionParameters}
\end{figure*}

Figure~\ref{ComparisionParameters} compares ICE vehicles, EVs, and HEVs in terms of physics, performance, and power consumption. ICE vehicles rely on high-energy-density fossil fuels, EVs use lower-energy-density batteries, and HEVs combine both systems. EVs provide instant torque and rapid acceleration, while ICE vehicles depend on crankshaft rotation \citep{malik2025energy}. HEVs improve range and efficiency by integrating both systems. In energy efficiency, EVs lead (85--90\%) due to regenerative braking and efficient motors, ICE vehicles are lowest (25--30\%) due to mechanical losses, and HEVs lie in between \citep{veza2023electric}. Maintenance is highest for ICE vehicles, lowest for EVs, and moderate for HEVs due to reduced mechanical stress \citep{burnham2021comprehensive}.

Existing research on vehicle power consumption estimation can be categorized into physical, statistical, and machine learning (ML) models. Physical models compute power based on dynamic parameters (\textit{e.g.,} vehicle speed \citep{shen2023personalized}, acceleration, rolling resistance \citep{madhusudhanan2021computationally}, traffic flow, road slope \citep{shen2022electric}, air density, and temperature) and static parameters (\textit{e.g.,} vehicle mass, frontal area, and drag coefficient \citep{madhusudhanan2021computationally}), applying fundamental physical laws \citep{zhang2024review}. These models are interpretable and accurate under standard conditions, but suffer limited adaptability to diverse driving behaviors and environments \citep{zhang2024review}. Moreover, they require many parameters that are difficult to obtain in real-time and pose challenges in development due to complex interdependencies that must preserve causality \citep{yu2024learning}.

On the other hand, statistical models use regression techniques such as linear, multiple, and non-linear regression to analyze the relationship between influencing factors and power consumption. These models often are derived from physical principles \citep{zhang2024review} and employ descriptive statistics to identify patterns and trends \citep{huang2024analysis}. Although they offer simplicity and computational efficiency, they struggle to capture the complex, nonlinear dynamics of modern powertrains (\textit{e.g.,} regenerative braking), especially in HEVs. Moreover, their performance is sensitive to noise and feature selection, making them less robust in dynamic driving environments \citep{zhang2024review}.

To address the limitations of physical and statistical models, ML-based approaches have been developed to predict power consumption by learning directly from data. These models can process large datasets and characterize complex patterns associated with energy use, making them particularly effective for capturing the intricate dynamics of modern vehicles with complex powertrains. ML models can be categorized into traditional models (\textit{e.g.,} support vector machines (SVM), random forest (RF), and gradient boosting machine (GBM)), and deep neural network (DNN) (\textit{e.g.,} convolutional neural networks (CNN), temporal convolutional networks (TCN), long short-term memory (LSTM) and Transformers). In scenarios involving large datasets and complex relationships between vehicular signals and power consumption, DNNs typically outperform traditional models due to their greater capacity to model non-linear and temporal dependencies, and extract high-level features.

Building on the capabilities of ML models, it becomes crucial to distinguish between different forms of power consumption. Understanding vehicle power usage requires not only analyzing how power fluctuates moment-to-moment but also how it accumulates over time. Although modeling both instantaneous and cumulative power consumption is essential for a comprehensive understanding of power efficiency and emissions of road vehicles, few studies address both aspects. Instantaneous power consumption models aid in route optimization, powertrain efficiency assessment, and evaluation of driver behavior, while cumulative power data supports long-term efficiency planning \citep{ansari2022estimating}. Predicting both enables more informed real-time decision-making and improves overall power efficiency \cite{chen2021data}.

However, even with robust modeling strategies, the reliability of power consumption predictions is influenced by the presence of uncertainty in real-world driving environments. The accuracy of power estimation models across different vehicle powertrains is affected by various sources of uncertainty. Factors such as driving style, traffic patterns, sensor noise, and parameter estimation inaccuracies introduce variability in predictions. While lab tests offer controlled conditions, they often fail to reflect real-world complexities, leading to discrepancies in fuel or power consumption estimates \citep{liu2021modelling}. These uncertainties are more significant in EVs and HEVs due to their intricate electric systems, including variability in battery capacity, regenerative braking, and nonlinear motor dynamics \citep{khiari2023uncertainty}. Although many studies have addressed vehicle power consumption estimation for ICE, EV, and HEV powertrains, a notable gap remains in the literature concerning the quantification and analysis of uncertainty in these estimations. Understanding uncertainty is essential alongside prediction accuracy, particularly given the complex behavior of EVs and HEVs. Accurate uncertainty quantification improves model reliability, robustness, and applicability in scenarios such as energy management, range estimation, and adaptive control. Overlooking uncertainty can introduce critical risks, especially for EVs and HEVs, where regenerative braking and hybrid mode transitions contribute additional variability. Despite its importance, uncertainty quantification for power estimation across all powertrain types is underexplored, marking a critical research gap. This is especially vital for DNNs, which are susceptible to overconfident predictions \citep{he2023survey}.

To address these challenges and gaps, our main contributions in this study are as follows:

\begin{itemize}
    \item We conducted a comprehensive data-driven approach to predict vehicle power consumption using DNNs, including TCN, LSTM, and Transformer, as well as RF as a traditional ML model. This approach provides a ML-based model for predicting the power consumption of any vehicle, regardless of the powertrain type (\textit{e.g.,} ICE, EV, or HEV). 
    \item To develop a reliable data-driven model for vehicle power prediction, access to accurate and representative real-world data is crucial. A key limitation in many existing studies is their reliance on simulated datasets, which often fail to capture the variability of actual road conditions. To address this gap, a key contribution of this study is the collection of comprehensive real-world driving data, including powertrain dynamic features, from three distinct powertrain types: ICE vehicles, EVs, and HEVs.
    \item Our study covered both instantaneous and cumulative power consumption to provide a complete picture of power behavior on timescales, which is vital to improving energy efficiency, vehicle control, and sustainability.
    \item We estimated and compared the uncertainty of different vehicle data and models using a frequentist Monte Carlo-based approach, by applying the model's weight random initialization, additive Gaussian noise, and dropout.
\end{itemize}

% In this study, we develop a data-driven approach using both traditional ML algorithms and DL models to estimate the power consumption of vehicles with different powertrains. Some studies refer to it as fuel consumption for ICE vehicles and energy consumption for EVs and HEVs. Since our study involves all three vehicle types, we use the term ``power'' consumption prediction to represent the appropriate consumption measure for each type. Specifically, we build models to estimate both instantaneous and cumulative power consumption and identify the most influential powertrain dynamic features for each powertrain type. 
The rest of this paper is organized as follows. Section~\ref{sec: related_work} provides a brief overview of the relevant literature and highlights existing gaps. In Sections \ref{Dataset}, the data acquisition and synchronizing procedures for three different vehicle types are reviewed. In Section~\ref{sec:methodology}, we describe the ML models and uncertainty approach employed. Section \ref{EXPERIMENTAL} presents the evaluation results across different vehicle types and feature sets. 
% We also compare the models using a shared subset of features among the three vehicle types to investigate how prediction accuracy varies depending on vehicle dynamics. Given the fundamental differences in powertrain architectures, ICE, EV, and HEV, we focus on fuel power estimation for ICEs, electric power for EVs, and a combination of both for HEVs. This allows for a fair and meaningful comparison across vehicle types. The hyperparameter tuning procedure is discussed as well. 
In Section~\ref{sec:discussion}, we compare our results with similar works in the literature. Finally, Section~\ref{sec:conclusion} summarizes our findings and outlines potential directions for future research.

\section{Related Work} \label{sec: related_work}
This section presents relevant power estimation methods for each type of vehicle.

\subsection{ICE Vehicles}

 % \cite{shahariar2023real} employed ML models to estimate fuel consumption and emissions (CO\textsubscript{2} and NO\textsubscript{x}) using real-world driving data from a light-duty diesel vehicle. The dataset, collected from 30 participants driving on an urban test route, included engine parameters (\textit{e.g.}, engine speed, load, throttle position) and driving dynamics (\textit{e.g.}, speed, acceleration, road type), using a portable emission measurement system to measure fuel consumption and emissions directly from exhaust. Three models were evaluated: linear regression, SVM, and Gaussian process regression.
 \cite{heni2023measuring} introduced a GBM model to predict vehicle fuel consumption. It used geospatial and temporal features derived from global positioning system position (GPS) and controller area network (CAN)-bus data, including speed, acceleration, and road gradient. The model was trained on real-world driving data collected from a light-duty vehicle in both urban and highway conditions. Recently, \cite{wang2023predictability} introduced an LSTM-based model optimized via Bayesian hyperparameter tuning for predicting vehicle fuel consumption. It used time series data including vehicle speed, acceleration, and road slope, collected from On-board diagnostics (OBD)-II and smartphone sensors.
 Moreover, \cite{ansari2022estimating} developed an ML model to predict instantaneous fuel consumption using real-time OBD data. They applied RF and neural network (NN) algorithms with engine load, speed, manifold pressure, air-fuel ratio, and throttle position as input features. Data were collected from a 100 \,km mixed driving route (urban, residential, highway) using a Ford Escape and a Ford F-350. Fuel consumption and OBD data were recorded at 5 and 2 \,Hz, respectively, using a Sentronics FlowSonic fuel flow meter and a CANedge2 data logger. 
 % \cite{kanarachos2019instantaneous} proposed a fuel consumption estimation method using the LSTM network on smartphone-based data, including global positioning system position (GPS), speed, altitude, acceleration, and the number of visible satellites, to estimate instantaneous fuel consumption.

\subsection{EVs} 
% A major advantage of EVs over ICE vehicles is their ability to recover energy through regenerative braking systems, which convert kinetic energy into electrical energy to recharge the battery \citep{wu2015electric}. Accurately accounting for regenerative braking is crucial when measuring EV energy consumption \cite{madhusudhanan2021computationally}. Predicting EV energy consumption is essential for improving efficiency and promoting adoption \cite{wu2015electric}. Accurate forecasts help reduce range anxiety and build driver confidence in EV technology \cite{shen2023personalized}. However, the widespread adoption of EVs introduces challenges, especially increased pressure on power grids caused by uncoordinated charging patterns. To address this, efficient charging schedules and algorithms are needed to manage demand and maintain grid stability \cite{rathore2023prediction}. 

\cite{nabi2023parametric} adopted a parametric analysis and an ML model to estimate the energy consumption in EVs.  Parametric analysis was based on a one-dimensional EV model using GT-Suite software to estimate energy consumption across standardized driving cycles. The model simulates parameters such as battery power, motor power, speed, and state of charge (SoC), allowing regenerative braking during deceleration. However, they did not use real driving data and instead relied on simulated data. In another study, \cite{shen2022electric} proposed a hybrid framework combining Transformer and Markov-Chain Monte Carlo (MCMC) models to predict EV's velocity and energy consumption. The model's input features were route factors, vehicle position, velocity, and driver behavior. Transformers predicted velocity changes near stops and turns, while MCMC models captured speed variations during normal driving. Data were collected from a Nissan Leaf SV over two weeks, recording speed, battery status, and GPS at 1\,Hz. 

In a different investigation, \cite{qi2018data} estimated cumulative energy consumption in EVs using positive and negative kinetic energy to represent acceleration energy and regenerative braking effects, respectively. A 4\textsuperscript{th}-order polynomial regression model was employed with average speed to capture nonlinear speed-energy relationships. The study also incorporated driving style indicators (\textit{e.g.}, speed, acceleration, deceleration), weather attributes (\textit{e.g.}, temperature, air pressure, wind speed), infrastructural factors (\textit{e.g.}, road type, slope, traffic calming), and traffic intensity as input features. Moreover, some studies have considered different parameters to estimate energy consumption in EVs, for example, SoC, battery power, motor power, slope of the road, and current of the battery \citep{nabi2023parametric}, \citep{achariyaviriya2023estimating}.

\subsection{HEV}

\cite{zeng2018modelling} proposed a non-linear regression model using NNs and least squares SVMs to estimate the energy consumption of an HEV. The model took transient motor torque, motor speed, and differential energy consumption as inputs to predict the DC bus current, and used the cumulative energy consumption and other variables to predict the DC bus voltage. In prior work, \cite{estrada2023deep} introduced a hybrid modeling approach to evaluate energy management strategies and estimate pollutant emissions in HEVs. The authors used experimental data collected from a chassis dynamometer test of a conventional turbocharged gasoline engine vehicle, using input features such as engine speed, air mass flow, torque, and exhaust temperature. CNNs were trained to estimate instantaneous and cumulative emissions of CO, NOx, THC, and CO$_2$, and were integrated with physical models of the rest of the powertrain in a MATLAB-Simulink and AMESIM co-simulation environment. Recently, \cite{zhang2023energy} developed an energy consumption model for HEVs using a minimum equivalent fuel consumption approach, which equated electric motor energy use with ICE fuel consumption. The model defined total energy consumption as the sum of fuel consumption and battery power converted into an equivalent fuel value. Inputs included battery and engine power, vehicle speed, acceleration, and road conditions, while outputs optimized energy distribution in real-time between the ICE and electric motor. Nonetheless, they overlooked cumulative energy consumption. \cite{jeong2024comparison} employed physical models to compare the energy consumption of EVs and HEVs using engine-specific parameters for HEVs, such as engine speed and engine torque, along with motor and battery parameters for both EVs and HEVs, including motor speed, motor torque, battery current, battery voltage, and SoC.

\subsection{Uncertainty analysis of vehicle power consumption}
Recent studies have adopted frequentist Monte Carlo techniques to quantify uncertainty in DNNs, often by introducing randomness during inference through multiple forward passes \citep{gal2016dropout, hashemipour2022uncertainty}. For instance, \cite{yagli2022ensemble} employed Monte Carlo dropout, performing numerous stochastic forward passes with dropout inference, to generate an ensemble of solar power forecasts without retraining multiple models. This approach ensures diverse network parameters on each pass, mimicking an ensemble and capturing the uncertainty of the model. Similarly, \cite{yalamanchi2023uncertainty} combined random Gaussian weight noise and dropout to form a Monte Carlo ensemble for DNN-based prediction of fuel properties. They showed that using multiple forward passes with different dropout rates or perturbed weights produces a spread of outputs reflecting the uncertainty of the prediction. This method satisfies the requirement for multiple sources of randomness through the use of dropout during inference, random initialization of network weights, and Gaussian noise added to inputs. These stochastic elements introduce diverse perturbations into the model’s behavior, enabling robust uncertainty estimation. Supported by recent literature, This Monte Carlo-based ensemble strategy, combining dropout \citep{gal2016dropout}, noise injection \citep{lakshminarayanan2017simple}, and weight variability \citep{kendall2017uncertainties}, provides a practical and validated approach for uncertainty quantification in data-driven vehicle power consumption modeling \citep{khiari2023uncertainty}.

\subsection{Research gaps}

Based on the literature, one of the most notable gaps is the lack of a comprehensive dataset that includes all three vehicle types: ICE, EV, and HEV. To address this, we collected a complete dataset including all vehicle types along with their powertrain dynamic features.

Estimating power consumption in EVs and HEVs presents significant challenges due to the complexity of their powertrain systems. First, estimating instantaneous power consumption in EVs and HEVs is challenging due to the high variance in signal distributions, leading to potential inaccuracies \citep{hussein2018adaptive}. To address this, we proposed a windowing approach, where a sliding window captures temporal dependencies in time series. This approach improves context awareness, enhances pattern recognition, reduces prediction errors, and mitigates noise, resulting in more accurate power estimation. Another major challenge is accounting for regenerated energy during deceleration, which many studies did not fully address due to its complexity \citep{shen2022electric, shen2023personalized}. Unlike conventional fuel consumption, EV power consumption is not directly correlated with speed, as energy can be consumed or regenerated at any speed. Neglecting this factor distorts estimates, particularly in urban driving with frequent acceleration and braking \citep{zhang2024review}. To address this, we incorporated acceleration as a key feature, recognizing that negative acceleration (deceleration) indicates regenerative braking (negative power), while positive acceleration signifies power consumption \citep{wu2015electric}.

Another significant gap in the literature is the analysis and comparison of uncertainty across different powertrains. To address this, we proposed a Monte Carlo frequentist approach to model, assess, and compare the uncertainty associated with each powertrain.

\section{Data Description and Preprocessing} \label{Dataset}
\begin{figure}[ht] 
\begin{center}
\centerline{\includegraphics[width=0.5\columnwidth]{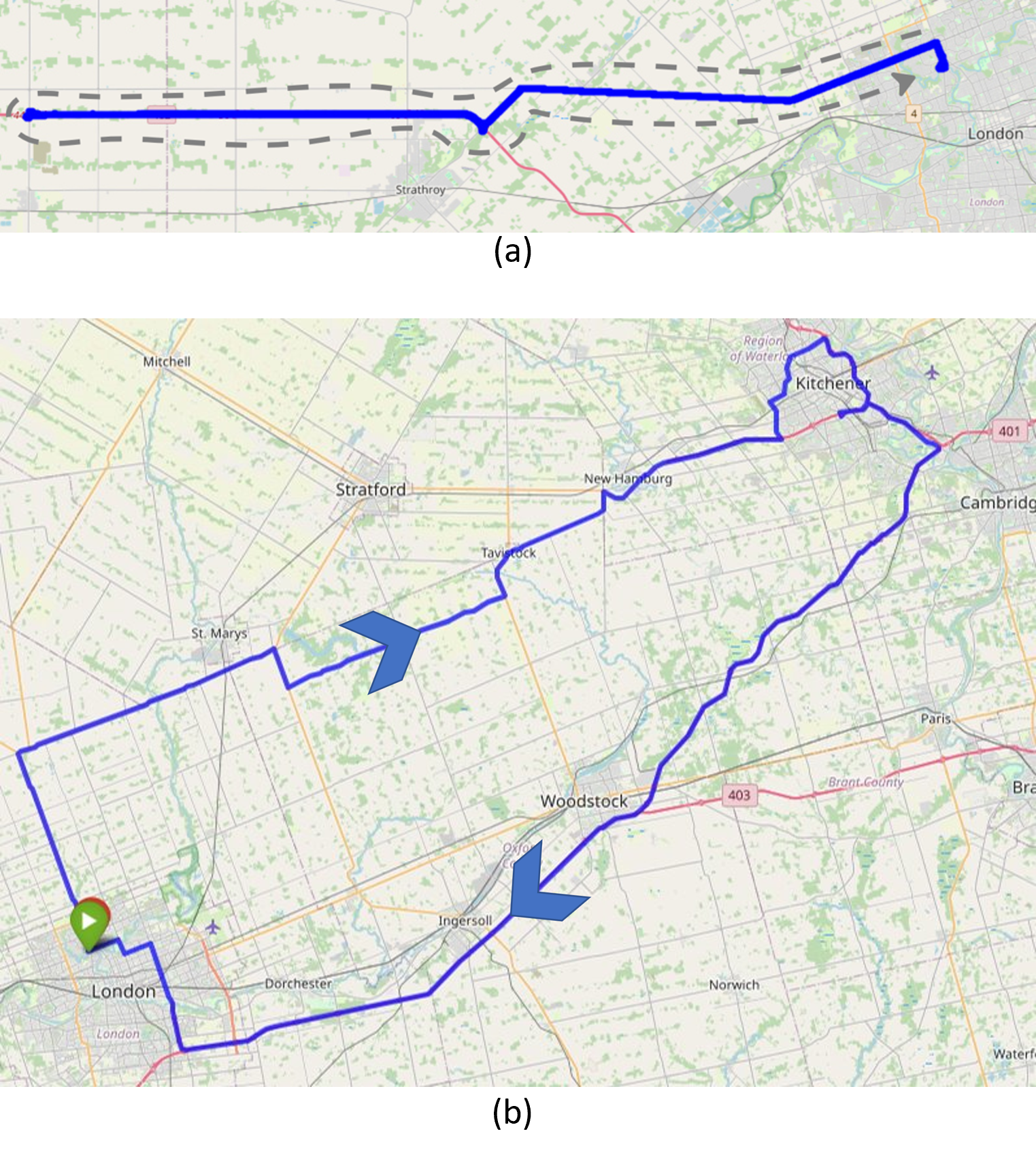}}
\caption{Rout maps for data collection in Ontario, (a) London to Sarnia, and (b) London to Waterloo.}
\label{Route}
\end{center}
\vskip -0.1in
\end{figure}

In order to conduct a comprehensive study, we collected real driving datasets from three types of vehicles: ICE, EV, and HEV. We collected approximately 12 hours of driving data, including city and highway routes (approximately 4 hours per powertrain type), on two different routes (see Fig. \ref{Route}). The first route extends from London, Ontario, to Sarnia, Ontario, covering both outbound and return trips. The route was selected to ensure that most of the trip is within the 403 Highway, where the speed limit is 110 \,km/h. The second was a closed route that started from London, Ontario, to Waterloo, Ontario, and back, where most of the trip was on Highway 7 with a typical speed limit of 80\,km/h. Table~\ref{samples} presents the number of samples in each dataset (ICE, EV, and HEV) for the training, validation, and testing sets, along with the sampling rate for each vehicle type.

\begin{figure}[] 
\begin{center}
\centerline{\includegraphics[width=0.4\columnwidth]{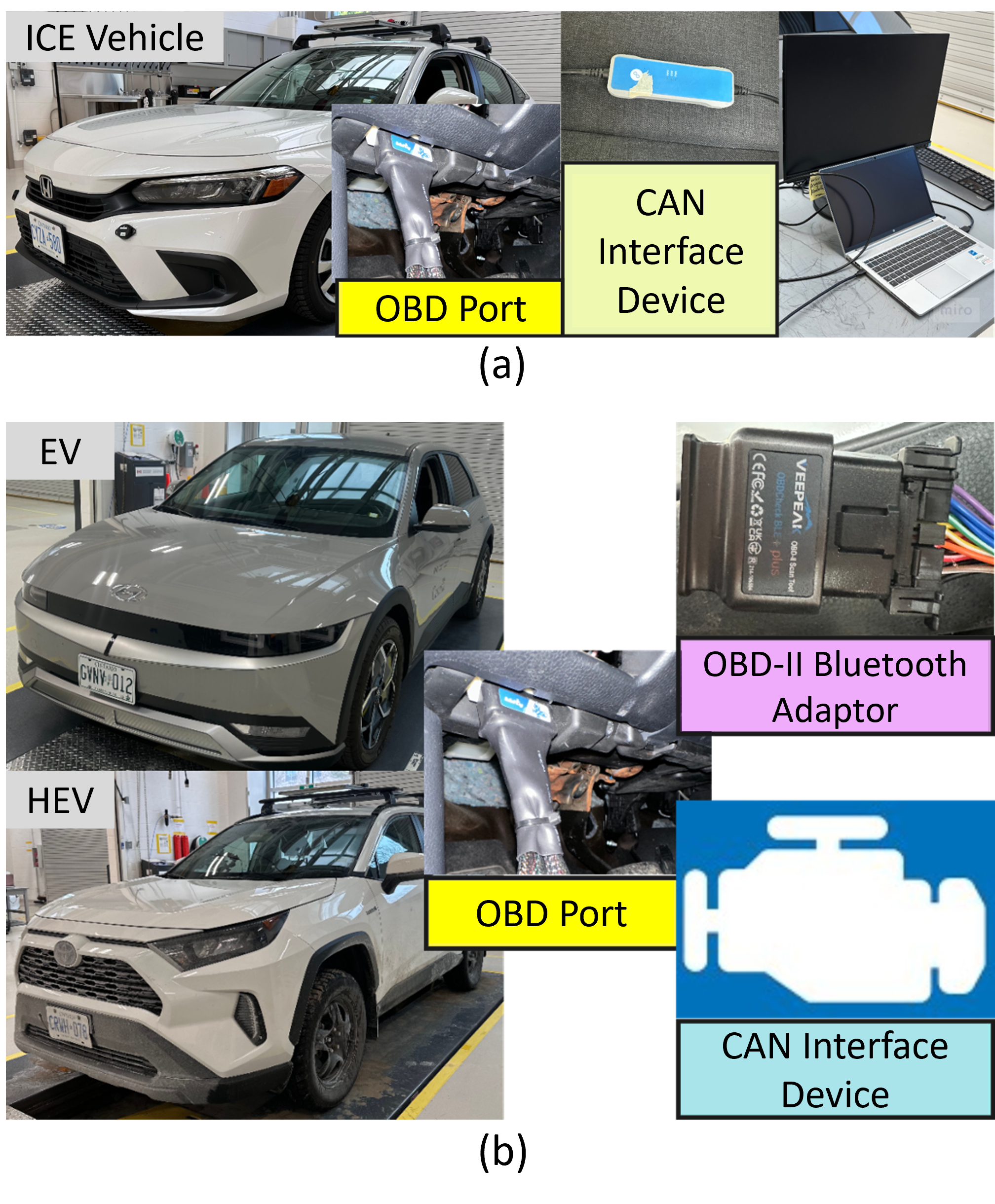}}
\caption{Data acquisition sensors for our (a) ICE vehicle and (b) EV and HEV.}
\label{ICE}
\end{center}
\vskip -0.1in
\end{figure}

The ICE data were collected via connecting a CAN-bus interface to the OBD port, connecting it to a laptop. For EV and HEV datasets, interpreting CAN-bus data is more complex due to security restrictions; therefore, we used the ``Car Scanner'' smartphone application connected to the OBD-II port via OBD-II Bluetooth adapter to the vehicle’s OBD port, pairing the app with the adapter, and scanning the car to retrieve the specified parameter IDs (PID). Since the sampling rate of the CAN-bus in the ICE vehicle differs from that of the smartphone application used for EV and HEV, the total number of samples varies across datasets. The data collection setup is illustrated in Fig.~\ref{ICE}, and Table~\ref{Input data} summarizes the dynamic signals collected for each powertrain type in terms of input and target data. Additionally, three battery-related parameters (SoC, battery voltage, and battery current) were recorded for comparison purposes only, as the proposed method relies solely on dynamic signals.

% \textcolor{blue}{Both routes should be shown here.}

\begin{table}[]
\centering
\caption{Number of samples in each collected dataset (ICE, EV, and HEV).}
\label{samples}
\begin{tabular}{@{}llllll@{}}
\toprule
Dataset & Training & Validation & Testing & Total & Frequency (Hz) \\ \midrule
ICE     & 21,000   & 3,000      & 6,000   & 30,000 & 2 \\
EV      & 2,275    & 325        & 650     & 3,250  & 0.2 \\
HEV     & 17,500   & 2,500      & 5,000   & 25,000 & 1.7 \\ \bottomrule
\end{tabular}
\end{table}

The ICE vehicle was a 2022 Honda Civic, for which both instantaneous and cumulative fuel power consumption were selected as target variables. The EV used in this study was a 2022 Hyundai Ioniq 5, and the HEV was a 2021 Toyota RAV4. The data acquisition procedure for the HEV followed the same approach as the EV. However, the HEV dataset included additional fuel-related parameters that distinguish it from the EV, such as engine torque, which complements the motor torque data available in both EV and HEV vehicles. 
% Also, the vehicle speed signal for each vehicle type is illustrated in Fig. \ref{SignalPlots}.

\begin{table}[t!]
\caption{The parameters of the collected data for ICE vehicle, EV and HEV.}
\small
\label{Input data}
\begin{center}
\begin{tabular}{@{}lllll@{}}
\toprule
Parameter              & Unit & ICE & EV & HEV \\ \midrule
\multicolumn{5}{c}{Input Data}                 \\ \midrule
speed          &   km/h   &  \checkmark   & \checkmark   & \checkmark    \\
acceleration   &   m/$s^2$   &   \checkmark  &  \checkmark  &  \checkmark   \\
engine torque          &   N.m   &   \checkmark  &    &  \  \\
motor torque           &   N.m   &     & \checkmark   &     \\
engine RPM             &   RPM   &   \checkmark  &    &  \checkmark   \\
motor RPM              &   RPM   &     & \checkmark   &     \\
 \hdashline
\multicolumn{5}{c}{Only for Comparison}                 \\ \hdashline
SoC        &  V      &     &  \checkmark  &  \checkmark   \\
Battery voltage        &  V      &     &  \checkmark  &  \checkmark   \\
Battery current        &   A     &     & \checkmark    &   \checkmark  \\ \midrule
\multicolumn{5}{c}{Target Data}                \\ \midrule
Instantaneous fuel power &   kW   &  \checkmark   &    &  \checkmark   \\
Instantaneous electric power     &   kW   &     & \checkmark   &  \checkmark   \\
Cumulative fuel power &   kW   &  \checkmark   &    & \checkmark    \\
Cumulative electric power &   kW   &     &  \checkmark  &   \checkmark  \\
\bottomrule
\end{tabular}
\end{center}
\end{table}

% \begin{figure}[!t] 
% \begin{center}
% \centerline{\includegraphics[width=0.6\columnwidth]{Pics/SignalPlots.png}}
% \caption{Visualization of the vehicle speed signal for (a) ICE, (b) EV, and (c) HEV vehicles.}
% \label{SignalPlots}
% \end{center}
% \vskip -0.1in
% \end{figure}

% \subsection{Data Preprocessing} \label{Preprocessing}
\subsection{Data synchronization}
Powertrain dynamic signals are recorded at different sampling frequencies due to signal prioritization, sensor sampling rates, and CAN-bus bandwidth limitations. High-priority signals, such as engine RPM and vehicle speed, require frequent updates for real-time monitoring, while lower-priority signals, such as air conditioning status, update less frequently. Additionally, each sensor and electronic control unit operates at a specific sampling rate depending on the nature of the data it collects. To ensure accurate data alignment for model training, signal synchronization is an essential preprocessing step. Since all signals were captured through a CAN-bus interface or app, they share a global timestamp. Synchronization was performed by selecting the frame rate of one signal as a reference and prioritizing the least noisy signal. For each frame and signal, the nearest timestamp to the reference timestamp was identified, and the corresponding value was assigned. To achieve precise synchronization, a timestamp resolution of $10^{-7}$ seconds was used, matching the precision of the global timestamp.

\subsection{Windowing approach} \label{Windowing approach}
To effectively capture temporal dependencies in vehicle power consumption, we employed a non-overlapping sliding-window approach that moves across the signals over time. 
% as shown in Fig. \ref{Windowing}. 
This method segments continuous time series data into sequences, ensuring that the model learns from both past and present states. Instead of processing individual data points in isolation, the network is fed structured sequences of input data, allowing it to recognize patterns, trends, and dependencies crucial to predicting instantaneous power consumption. The window size, which determines how much historical data is included in each sequence, is a hyperparameter that requires optimization to balance the capture of long-term dependencies and the maintenance of computational efficiency (as we found the optimized value in Subsection \ref{Hyperparameter}). A larger window size enabled the model to account for extended temporal relationships, while a smaller window focused on short-term variations. Since the data were time series, it was split chronologically to preserve temporal dependencies: the earliest portion was used for training, followed by subsequent portions for validation and testing, respectively.

% \begin{figure}[H] 
% \begin{center}
% \centerline{\includegraphics[width=0.6\columnwidth]{Pics/Windowing.png}}
% \caption{Windowing approach and data splitting.}
% \label{Windowing}
% \end{center}
% \vskip -0.1in
% \end{figure}

\section{Methodology} \label{sec:methodology}
\begin{figure*}[!t]
\centering
\includegraphics[width=\linewidth]{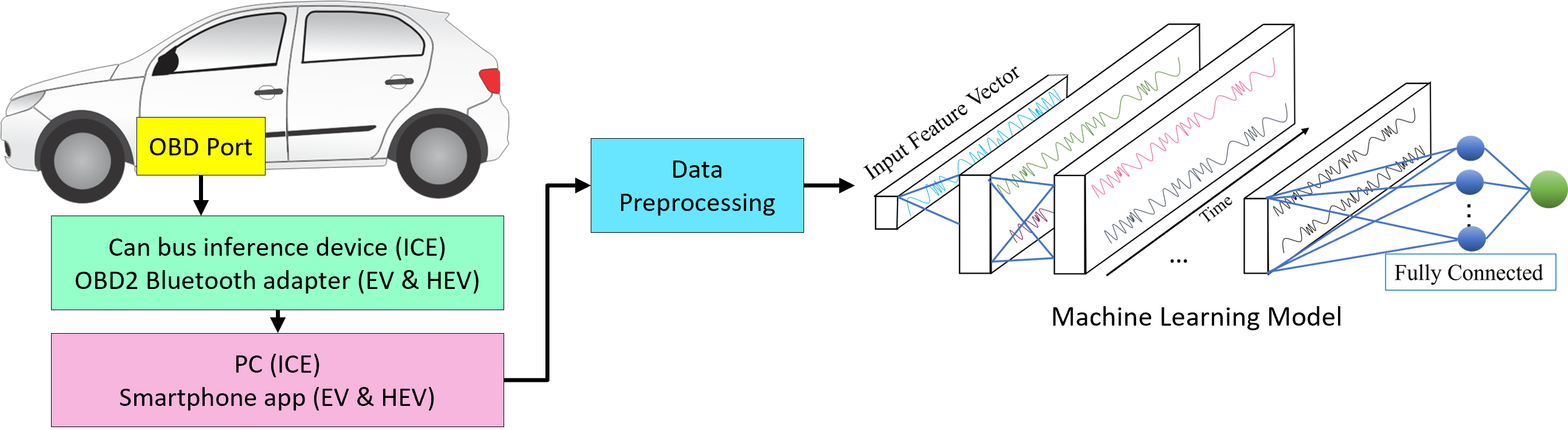}
\vskip -0.1in
\caption{Overview of the proposed framework: Dynamic data were acquired via CAN-bus or OBD-II interface, followed by preprocessing. Different feature sets were then provided to the ML models.}
\label{Framework}
\end{figure*}

We proposed a comprehensive data-driven approach for predicting power consumption in ICE, EVs, and HEVs powertrains. As depicted in Fig.~\ref{Framework}, our method covered the entire process, from data acquisition and preprocessing to model training and evaluation, ensuring an end-to-end solution for accurate power consumption prediction.

\subsection{Machine learning modeling}
In the context of vehicle power consumption prediction, input signals vary mainly over time. Therefore, to keep the temporal feature and time dependency, we have proposed three DNNs for time series analysis, and one traditional ML model, RF, as explained in the following subsections. 

\subsubsection{LSTM}
LSTM networks have been widely used to predict fuel, energy, and power consumption due to their ability to model complex temporal dependencies in sequential data. As an advanced type of recurrent neural networks (RNNs), LSTMs effectively mitigate the vanishing gradient problem, which often hampers traditional RNNs when learning from long sequences. LSTMs perform by incorporating a cell state and a set of gates (input, forget, and output) that regulate information flow, allowing the network to selectively retain or discard past information. This structure ensures that critical long-term dependencies, such as variations in driving cycles, regenerative braking effects, and acceleration patterns, are effectively captured. 

\begin{figure}[t]
  \centering
  \includegraphics[width=0.6\columnwidth]{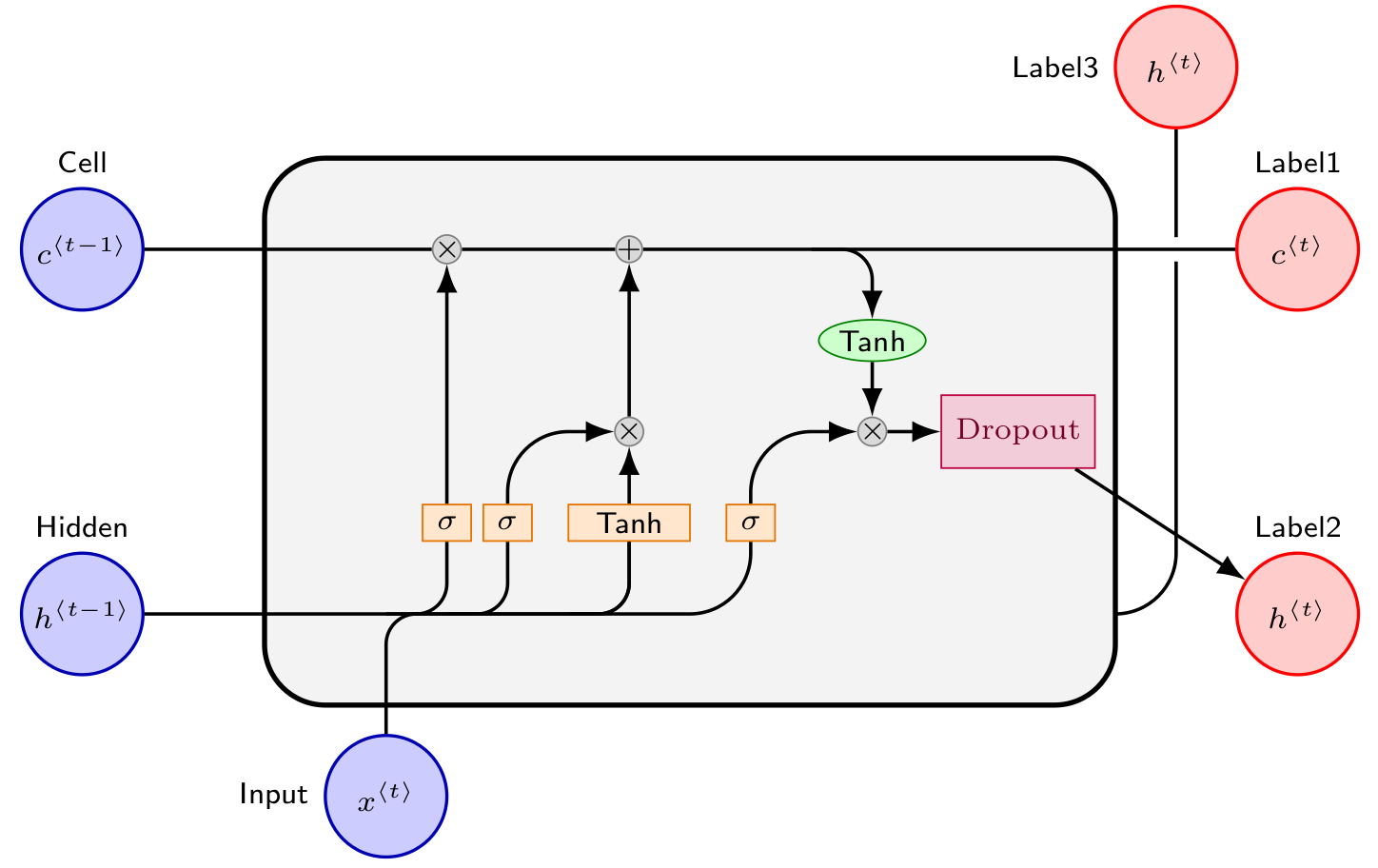}
  \caption{Architecture of the proposed LSTM} \label{LSTMArchitecture}
\end{figure}

As illustrated in Fig.~\ref{LSTMArchitecture}, the LSTM model processes an input sequence \( x^{\langle t \rangle} \) at each time step, along with the previous cell state \( c^{\langle t-1 \rangle} \) and hidden state \( h^{\langle t-1 \rangle} \), which encode the model’s long-term and short-term memory, respectively. The LSTM unit updates these to the new states \( c^{\langle t \rangle} \) and \( h^{\langle t \rangle} \) through its gated recurrent structure. In our implementation, the LSTM architecture consists of three stacked layers, each with 32 hidden units and a dropout probability of 0.2 to prevent overfitting. The model is configured with \texttt{batch\_first=True}, and only the final hidden state from the last time step is used for prediction. This output is passed through a dropout layer followed by a fully connected regression head that maps the hidden representation to a single scalar value representing the predicted power. This final output corresponds to the node labeled \textit{Predicted Power} in the diagram.
 
% The dropout layer is added in uncertainty analysis part.sa

\subsubsection{TCN}
TCNs are a class of NNs designed for sequential data processing, making them well-suited for time series prediction tasks \citep{lea2017temporal}. Unlike  RNNs, which process data sequentially, TCNs leverage convolutional layers with dilated convolutions to efficiently capture both short-term and long-range dependencies in the data. This architecture enables parallel processing, reducing computational complexity while maintaining the ability to model extended temporal relationships. We proposed TCN for power consumption estimation to assess its effectiveness in capturing intricate temporal patterns in vehicle dynamics. By applying dilated convolutions, the model expands its receptive field exponentially, allowing it to process long sequences without the vanishing gradient issues commonly associated with traditional RNNs. This makes TCN a promising alternative for modeling instantaneous and cumulative power consumption, as it can effectively learn complex dependencies across multiple time steps while maintaining stability during training.

\begin{figure*}[!t]
\centering
\includegraphics[width=\linewidth]{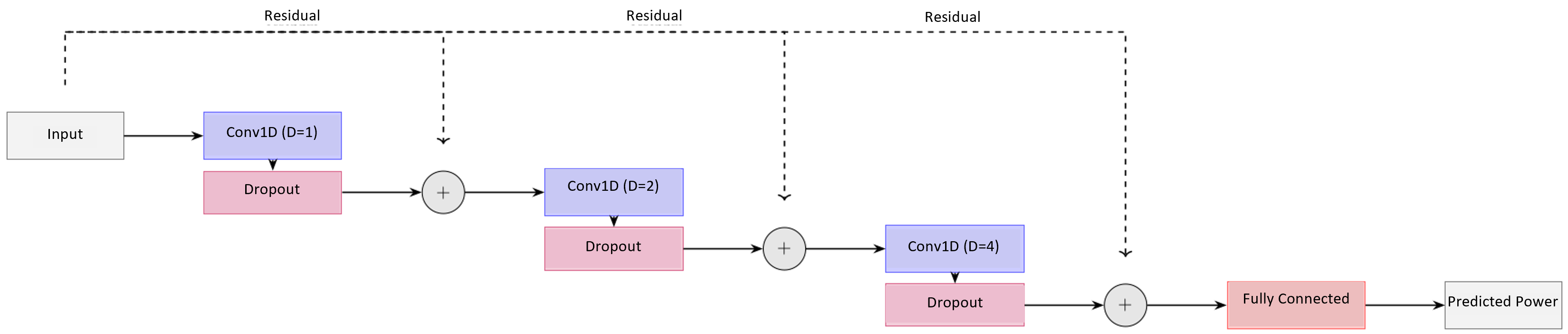}
\caption{Architecture of the proposed TCN.}
\label{TCNArchitecture}
\end{figure*}

The TCN model's architecture is illustrated in Fig. \ref{TCNArchitecture}. It consists of a stack of three 1D convolutional layers with increasing dilation factors (1, 2, and 4), each having a kernel size of 5 and followed by a dropout layer with a rate of 0.002 to prevent overfitting. These convolutional blocks capture temporal dependencies at multiple scales across the input sequence \( x \in \mathbb{R}^{T \times C} \), where \( T \) is the sequence length and \( C \) is the number of input channels. After passing through the TCN module, the output is transposed and the feature corresponding to the last time step is fed into a fully connected layer to produce the final output \( \hat{y} \in \mathbb{R} \). The diagram visualizes this architecture sequentially from left to right, highlighting the convolutional layers, dropout blocks, and the final prediction layer. 

\subsubsection{Transformer}
Transformers have been known as a powerful DNN architecture for time series forecasting and sequential data processing. Unlike recurrent models, Transformers utilize self-attention mechanisms to model long-range dependencies efficiently while enabling parallel processing. The architecture consists of several key components. The input sequence is first passed through a one-dimensional convolutional layer (\texttt{Conv1D}) to enhance feature extraction and increase the expressiveness of the input representation. The processed sequence is then encoded with positional information using a \texttt{PositionalEncoding} module, which applies sinusoidal and cosine functions to retain order information, addressing the lack of inherent sequential structure in Transformers. The core of the model consists of a multi-layer Transformer encoder, where each layer contains multi-head self-attention and feedforward layers. The self-attention mechanism enables the model to dynamically weigh the importance of different time steps in a sequence, allowing it to focus on crucial driving patterns. The processed sequence is then passed through a fully connected (\texttt{linear}) layer to generate the final instantaneous power prediction. The final prediction is based on the last time step of the sequence, ensuring that the model takes advantage of all relevant past information.

\begin{figure}[H] 
\begin{center}
\centerline{\includegraphics[width=0.4\columnwidth]{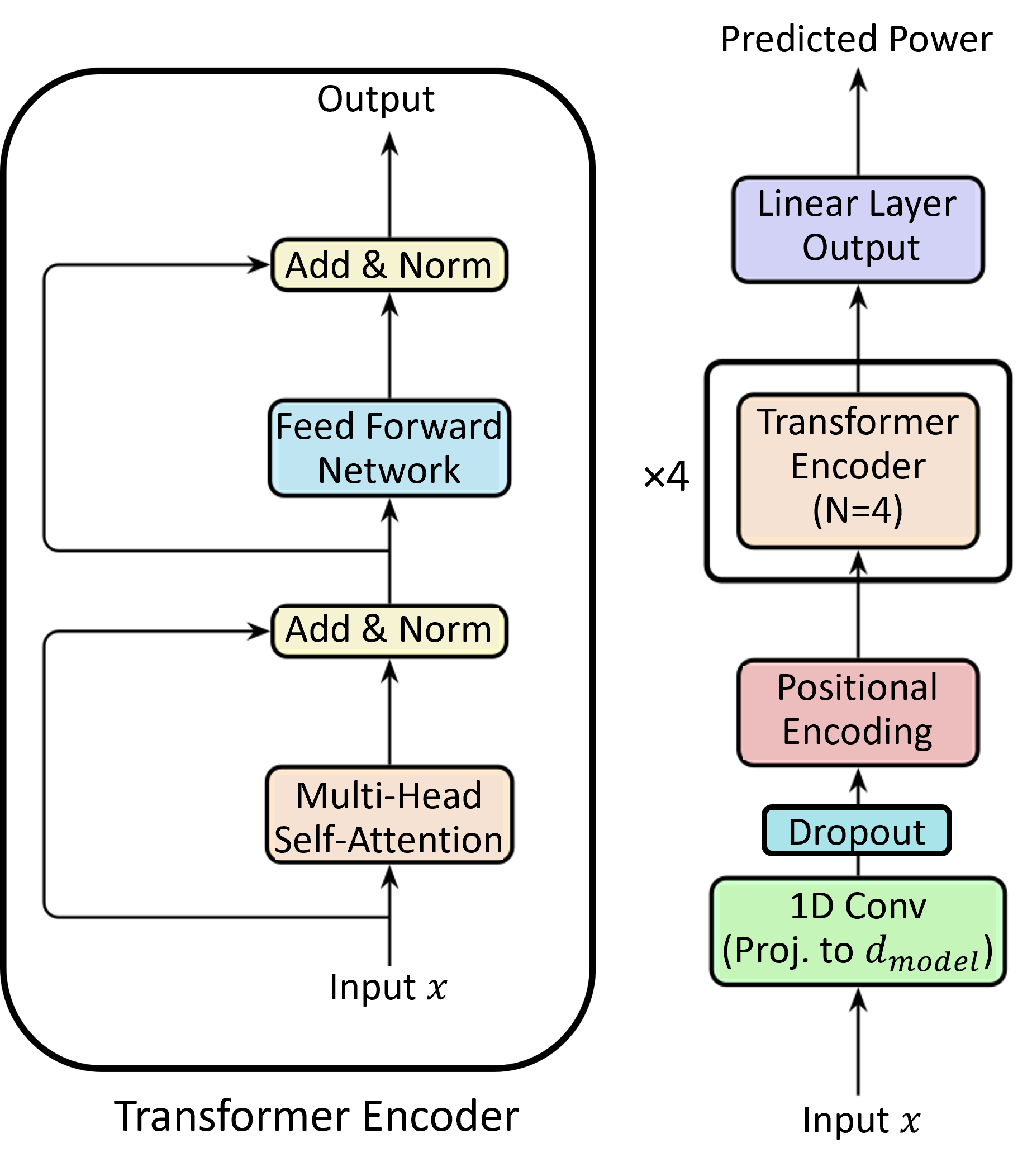}}
\caption{Architecture of the proposed Transformer.}
\label{TransformerArchitecture}
\end{center}
\end{figure}

As depicted in Fig.~\ref{TransformerArchitecture}, the Transformer architecture begins with a 1D convolution layer that projects the input time series signal into a higher-dimensional representation with \( d_{\text{model}} = 64 \) channels. A dropout layer follows the convolution. The output is then enhanced with ``Positional Encoding" block to retain temporal order. Then, the enriched sequence is passed through a stack of four Transformer encoder layers, each consisting of multi-head self-attention, feed-forward networks, and residual connections with layer normalization, abstracted as the Transformer Encoder (N=4) block in the diagram. Finally, the output corresponding to the last time step is passed through a fully connected linear layer to produce the final scalar prediction, as shown in the top ``Linear Layer" block leading to the output node.

\subsubsection{Random Forest}
RF is a widely used ensemble learning method that constructs a multitude of decision trees during training and outputs the average prediction of individual trees for regression tasks. Its inherent bootstrapping mechanism, where each tree is trained on a randomly sampled subset of the data with replacement, introduces natural variability in the model outputs, making it well-suited for capturing predictive uncertainty. Moreover, since each decision tree explores a different subset of features and samples, the aggregation of diverse predictions helps improve robustness and generalization. Due to its simplicity, interpretability, and built-in stochasticity, RF serves as a reliable baseline model for power consumption estimation and uncertainty quantification in regression problems. Additionally, feature importance is extracted directly from the trained ensemble, providing insights into how each input feature contributes to the prediction task. The limitation of RF in the power consumption prediction task lies in its inability to capture temporal dependencies, which are crucial when dealing with time series input data.

\subsection{Uncertainty analysis}
In this study, we quantified the uncertainty of power consumption prediction for ICE, EV, and HEV vehicle datasets using a frequentist Monte Carlo-based approach. To capture the uncertainty behavior of the input features, we modeled the Gaussian noise estimated from input data. \citep{hashemipour2022uncertainty}. For this purpose, we first calculated the feature-based Gaussian noise 
$\sim \mathcal{N}_{\mathrm{F}}(0, \sigma^2)$, 
where $\mathcal{N}_{\mathrm{F}}$ denotes a Gaussian distribution corresponding to feature $F$, with mean $0$ and variance $\sigma^2$. Then, $\mathcal{N}_{\mathrm{F}}$ was injected into training set. In addition, to follow the frequentist Monte Carlo approach proposed by \cite{gal2016dropout}, we repeated the forward process \(N\) times for each model while: 1- we injected the $\mathcal{N}_{\mathrm{F}}(0, \sigma^2)$ into test samples. 2- Dropout was activated in inference phase. 3- Random initialization of model weights in each training run. 

\subsection{Evaluation metric and protocol}
To evaluate time series forecasting models, common metrics include mean absolute error (MAE) and root mean square error (RMSE). The MAE is defined as \( \text{MAE} = \frac{1}{n} \sum_{i=1}^{n} \left| y_i - \hat{y}_i \right| \), and the RMSE is defined as \( \text{RMSE} = \sqrt{\frac{1}{n} \sum_{i=1}^{n} (y_i - \hat{y}_i)^2} \), where \( y_i \) is the actual power value, \( \hat{y}_i \) is the predicted value, and \( n \) denotes the total number of samples.

As mentioned in Section \ref{Introduction}, predicting both instantaneous and cumulative power consumption is essential for ICE vehicles, EVs, and HEVs to provide a comprehensive understanding of energy usage. Therefore, we analyze the power consumption instantaneously and cumulatively as follows.
\begin{equation}
    P_{\text{cumulative}}(t) = \int_{0}^{t} P_{\text{instant}}(t') \, dt',
\end{equation}
where \( P_{\text{instant}}(t) \) represents the instantaneous power consumption at time \( t \), and \( P_{\text{cumulative}}(t) \) denotes the cumulative power consumption up to time \( t \).

\noindent In the discrete case, the cumulative power consumption is approximated as follows.

\begin{equation}
    P_{\text{cumulative}}(t) = \sum_{i=1}^{n} P_{\text{instant}}(t_i) \cdot \Delta t,
\end{equation}

\noindent 
where \( \Delta t \) refers to the time interval between consecutive power measurements, and \( n \) indicates the total number of time steps.

% In this section, we conducted a comprehensive ablation study on the estimation of power consumption using four models, TCN, LSTM, Transformer, and RF in three types of vehicles, ICE, EV and HEV. The study aims to analyze two key aspects: first, how power consumption can be predicted using dynamic signals and which features are the most informative for each vehicle type; and second, how the power consumption prediction varies across different vehicle types.

\subsection{Hyperparameter tuning} \label{Hyperparameter}
We applied hyperparameter tuning to obtain the optimal hyperparameters for each model, using 10\% of the data as the validation set. For the hyperparameter optimization process, we chose the \textit{Optuna} method \citep{akiba2019optuna}. \textit{Optuna} offers several advantages for hyperparameter tuning in time-dependent DNNs. It uses Bayesian optimization via tree-structured parameter estimators enables efficient exploration of complex search spaces, which is particularly beneficial when dealing with the multiple interacting parameters found in sequential models. \textit{Optuna} also supports automatic pruning of unpromising trials during training, helping to reduce computation time by stopping less effective configurations early. This is an important benefit, given that longer training times are typical for temporal models. 
% Table \ref{Abb} shows the abbreviation of those hyperparameters considered for the tuning process in DNNs (\textit{e.g.,} TCN, LSTM, and Transformer).

% \begin{table}[!t]
% \caption{Abbreviation of optimized hyperparameters of DNNs.}\label{Abb}
% \centering
% \begin{tabular}{@{}ll@{}}
% \toprule
% Hyperparameter               & Abbreviations \\ \midrule
% Window size                        & WS                                \\
% Learning rate                      & LR                                \\
% Hidden dimension                   & HD                                \\
% Number of channels                 & NCH                               \\
% Number of layers                   & NL                                \\
% Kernel size                        & KS                                \\
% Number of heads                    & MH                                \\
% Dimension of embedding space       & ED                                \\
% Feedforward dimension              & FFD                               \\
% Dropout                            & DO                                \\ \bottomrule
% \end{tabular}
% \end{table}

% In the hyperparameter tuning process, the best-performing DNN model was selected for each dataset. 
Table \ref{HPO} summarizes the input characteristics and the HPs selected for the dataset for each vehicle type. Also, both MAE and RMSE were used as evaluation metrics to ensure consistent model performance. For each model and dataset, we performed separate tuning for both metrics, MAE and RMSE, and then evaluated the final models using the corresponding best configurations. The results showed that for all three models in the ICE, EV, and HEV datasets, the MAE optimized configuration consistently delivered superior performance. Specifically, for the ICE dataset, the best performing TCN model used WS = 10, LR = 0.01, HD = 64, NL = 3, and KS = 3. For the EV dataset, the Transformer achieved optimal results with WS = 50, LR = 0.001, ED = 64, MH = 2, NL = 2, FFD = 32, and DO = 0.1. For the HEV dataset, the LSTM model performed best with WS = 10, LR = 0.01, HD = 132, NL = 9, and KS = 5. Also for RF, the hyperparameter tuning was performed on ICE dataset. The optimized parameters include the number of estimators (\texttt{n\_estimators}) set to 100, which determines the number of decision trees in the forest, and the maximum depth of each tree (\texttt{max\_depth}) set to 20, which controls the depth to which each tree is allowed to grow to avoid overfitting. Additionally, an LR of 0.001 was used.

\begin{table*}[!t]
\caption{Optimized hyperparameters and their abbreviations: WS (Window Size), LR (Learning Rate), HD (Hidden Dimension), NCH (Number of Channels), NL (Number of Layers), KS (Kernel Size), MH (Number of Heads), ED (Embedding Dimension), FFD (Feedforward Dimension), and DO (Dropout).}
\label{HPO}
\centering
\resizebox{\textwidth}{!}{%
\begin{tabular}{@{}lll l@{}}
\toprule
Vehicle & Model & Hyperparameters & \begin{tabular}[c]{@{}l@{}}Optimization Metric\\ (MAE or RMSE)\end{tabular} \\ \midrule
ICE  & TCN         & WS, LR, HD, NL, KS                & MAE   \\
EV   & Transformer & WS, LR, MH, ED, NL, KS, FFD, DO   & MAE   \\
HEV  & LSTM        & WS, LR, HD, NL, KS                & RMSE  \\ \bottomrule
\end{tabular}
}
\end{table*}

\section{Experimental Results} \label{EXPERIMENTAL}
Our implementations were carried out on a 12\textsuperscript{th} Gen Intel\textsuperscript{\textregistered} Core\textsuperscript{\texttrademark} i7-12700 processor, 2.10 GHz, supported by 32.0 GB RAM and a 12.0 GB NVIDIA GeForce RTX 4090 GPU. The DNN implementation was performed using Pytorch.

\subsection{Implementation details}
We adopted three DNNs and one traditional ML model to predict instantaneous and cumulative power consumption. 
For all implementations, we normalized both training and target variable using the \texttt{MinMaxScaler} to ensure stable training. A sliding window approach was applied, where sequences of length \( 10 \) were created to capture temporal dependencies in the data. Also mean square error was used as the loss function for all models. 
% As can be seen in Fig. \ref{Windowing}, 
Dataset was split into training, validation, and testing sets (\( 70\% \), \( 10\% \), and \( 20\% \), respectively), and the training was performed in batches of 64 samples. A learning rate of 0.001 and Adam as the Optimizer were chosen. All models were implemented using PyTorch and for 300 epochs.

\subsection{ICE results}

\begin{table*}[!t]
% \scriptsize
\caption{Comparison of power consumption prediction using TCN, LSTM, Transformer, and RF models with different feature sets for ICE dataset.}
\label{ICEResults}
\centering
\resizebox{\textwidth}{!}{%
\begin{tabular}{@{}lllllllll@{}}
\toprule
 & \multicolumn{4}{c}{Instant Power Consumption Prediction} & \multicolumn{4}{c}{Cumulative Power Consumption Prediction} \\ \cmidrule(l){2-9} 
 & TCN & LSTM & Transformer & RF & TCN & LSTM & Transformer & RF \\ \midrule
 & \multicolumn{4}{c}{MAE$|$RMSE ($\times10^{-3}$)} & \multicolumn{4}{c}{MAE\%$|$RMSE\%} \\ \midrule
Speed & 2.0$|$3.4 & 1.7$|$3.3 & 2.0$|$3.9 & 3.4$|$5.2 & 13.70$|$15.82 & 16.07$|$19.10 & 33.67$|$37.89 & 33.12$|$37.7 \\
Acceleration & 5.8$|$6.7 & 5.8$|$6.8 & 7.7$|$8.0 & 6.6$|$7.3 & 40.15$|$54.28 & 30.49$|$42.90 & 33.73$|$47.45 & 45.87$|$58.93 \\
Engine RPM & 1.6$|$3.6 & 1.7$|$3.8 & 1.9$|$4.2 & 1.9$|$3.9 & 10.63$|$11.55 & 3.67$|$4.10 & 4.27$|$5.17 & 13.96$|$17.10 \\
Engine torque & 0.7$|$1.4 & 0.7$|$1.6 & 0.9$|$1.5 & 1.1$|$2.3 & 4.40$|$5.07 & 6.76$|$7.78 & 8.50$|$9.57 & 8.50$|$9.46 \\
{[}Speed, engine RPM{]} & 1.3$|$3.1 & 0.6$|$1.4 & 3.6$|$2.1 & 1.9$|$4.3 & 4.51$|$6.94 & 5.97$|$9.16 & 7.83$|$11.37 & 19.09$|$24.19 \\
{[}Speed, acceleration, engine torque{]} & 0.5$|$1.1 & 0.6$|$1.4 & 1.1$|$1.4 & 0.9$|$2.0 & 1.16$|$1.34 & 4.92$|$5.53 & 8.91$|$9.92 & 2.76$|$2.87 \\
{[}Speed, engine torque, engine RPM{]} & 0.4$|$0.9 & 0.6$|$1.3 & 1.4$|$1.9 & 0.7$|$1.5 & 1.02$|$1.07 & 1.22$|$1.48 & 1.82$|$2.13 & 3.16$|$3.71 \\
{[}Acceleration, speed, engine torque, engine RPM{]} & 0.5$|$0.9 & 0.6$|$0.9 & 1.1$|$1.3 & 0.7$|$1.5 & 2.41$|$2.88 & 1.32$|$1.59 & 1.40$|$1.67 & 3.05$|$3.58 \\ \bottomrule
\end{tabular}
}
\end{table*}

Table \ref{ICEResults}, shows the MAE and RMSE of four models' power consumption predictions on our ICE dataset for different feature sets. Based on the results presented in Table~\ref{ICEResults}, the performance of the models in the ICE dataset varies depending on the set of characteristics used. For instantaneous power consumption, the best performance is achieved with the set of features \([ \text{speed, engine torque, engine RPM} ]\) using the TCN model, which produces the lowest MAE (\(0.4 \times 10^{-3}\)) and RMSE (\(0.9 \times 10^{-3}\)). The LSTM model with the feature set \([ \text{acceleration, speed, engine torque, engine RPM} ]\) performs closely, achieving an MAE of \(0.6 \times 10^{-3}\) and an RMSE of \(0.9 \times 10^{-3}\). For cumulative power consumption, the TCN model with the \([ \text{speed, engine torque, engine RPM} ]\) feature set provides the best results with an MAE of \(1.02\%\) and RMSE of \(1.07\%\).

\begin{figure}[!t]
  \centering
  \includegraphics[width=0.5\columnwidth]{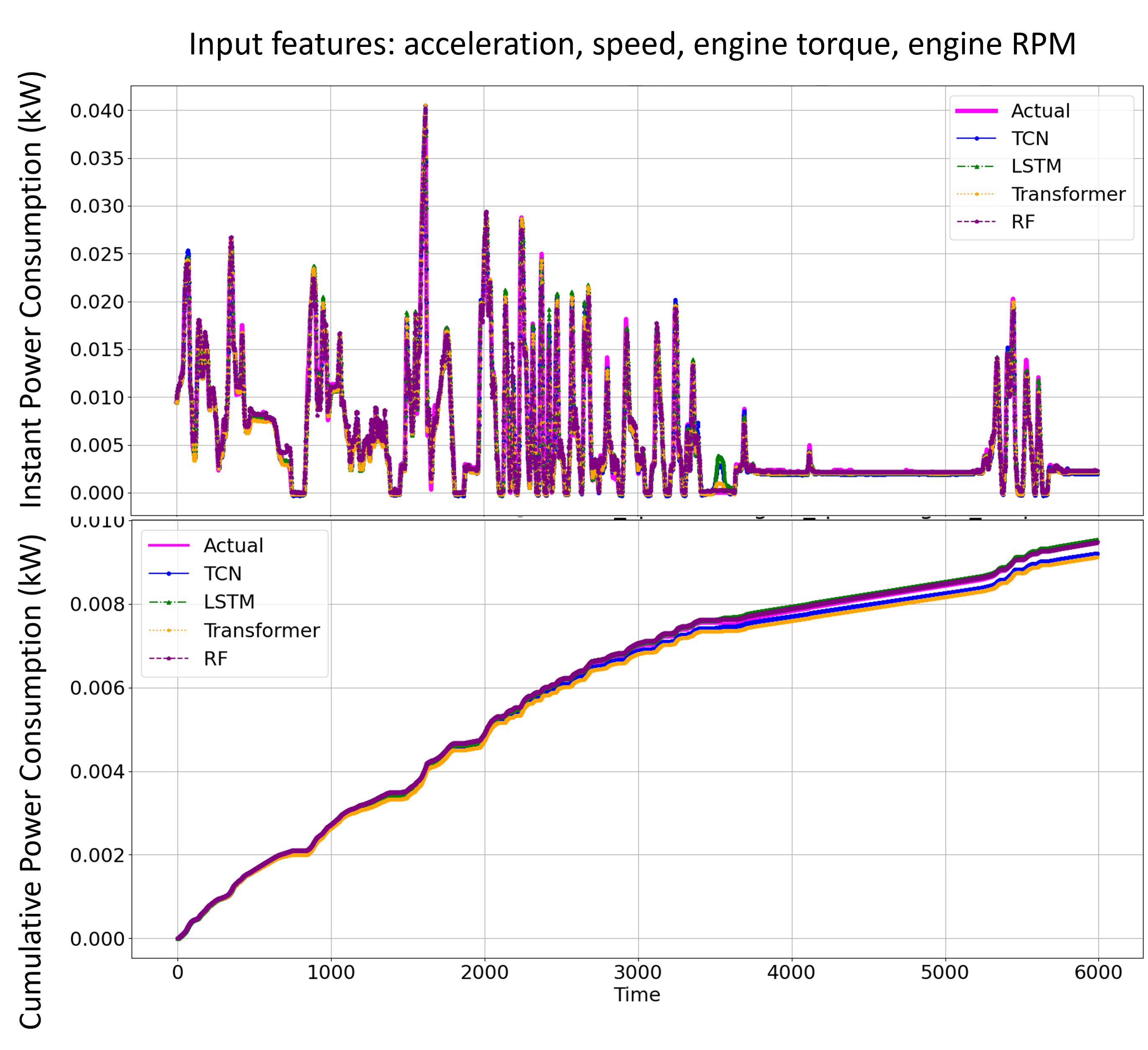}
  \caption{Actual and predicted instantaneous (upper) and cumulative (lower) power consumption in ICE dataset using the best performing (a) TCN, (b) LSTM, and (c) Transformer models.} \label{ICE_Plots}
\end{figure}

In terms of RF, while the result indicates the ability of RF to learn the data pattern, it generally performs worse compared to DNN models, especially for the prediction of cumulative power. For example, using the feature set \([ \text{acceleration, speed, engine torque, engine RPM} ]\), RF yields an MAE of \(0.7 \times 10^{-3}\) and an RMSE of \(1.5 \times 10^{-3}\) for instantaneous prediction, and MAE and RMSE percentages for cumulative power (MAE = 3.05\%, RMSE = 3.58\%) are much higher than those of DNNs. For instantaneous power prediction, DNNs, particularly TCN and LSTM, offer better performance than RF, particularly when using feature sets that include engine torque and speed. For cumulative power consumption, the TCN and LSTM models consistently provide more accurate predictions compared to the RF model, as they are capable of capturing temporal dependencies, which the RF model cannot. Figure~\ref{ICE_Plots} illustrates the predicted versus actual values for both instantaneous and cumulative power consumption using the feature set [acceleration, speed, engine torque, engine RPM] across four models.

\subsection{EV results}

\begin{table*}[!t]
\centering 
\caption{Comparison of power consumption prediction using TCN, LSTM, Transformer, and RF models with different feature sets for EV dataset.}
\begin{threeparttable}
\resizebox{\textwidth}{!}{%
\begin{tabular}{@{}lllllllll@{}}
\toprule
 & \multicolumn{4}{c}{Instant Power Consumption Prediction} & \multicolumn{4}{c}{Cumulative Power Consumption Prediction} \\ \cmidrule(l){2-9} 
 & TCN & LSTM & Transformer & RF & TCN & LSTM & Transformer & RF \\ \midrule
 & \multicolumn{4}{c}{MAE$|$RMSE} & \multicolumn{4}{c}{MAE\%$|$RMSE\%} \\ \midrule
Acceleration & 11.03$|$17.23 & 9.59$|$14.90 & 9.92$|$15.56 & 11.99$|$18.33 & 5.72$|$6.67 & 5.94$|$7.14 & 4.79$|$5.99 & 22.80$|$15.16 \\
Speed & 10.06$|$16.60 & 9.00$|$15.51 & 9.28$|$15.15 & 12.34$|$19.21 & 15.86$|$18.75 & 6.96$|$7.87 & 4.55$|$5.21 & 7.65$|$9.32 \\
Motor torque & 8.50$|$16.57 & 8.74$|$17.19 & 8.63$|$14.65 & 9.74$|$17.64 & 13.43$|$17.50 & 5.62$|$5.98 & 19.26$|$24.55 & 6.06$|$6.78 \\
Motor RPM & 10.55$|$13.38 & 7.83$|$14.45 & 8.67$|$13.92 & 12.07$|$19.05 & 2.75$|$4.17 & 6.62$|$8.00 & 7.24$|$8.76 & 9.59$|$11.01 \\
{[}Acc.\tnote{1} , speed{]} & 9.33$|$15.34 & 9.17$|$14.44 & 9.11$|$14.93 & 10.10$|$16.56 & 9.95$|$11.71 & 5.41$|$6.51 & 13.11$|$16.57 & 10.70$|$12.42 \\
{[}Acc.\tnote{1} , motor torque{]} & 8.77$|$17.21 & 8.42$|$16.59 & 8.06$|$14.31 & 8.33$|$14.90 & 4.03$|$4.64 & 7.16$|$9.68 & 22.74$|$22.61 & 8.85$|$9.83 \\
{[}Acc.\tnote{1} , motor RPM{]} & 10.98$|$18.75 & 8.94$|$14.22 & 10.36$|$17.20 & 10.70$|$17.45 & 7.57$|$9.75 & 4.82$|$5.70 & 19.46$|$24.29 & 13.51$|$16.02 \\
{[}Acc.\tnote{1} , speed, motor torque{]} & 5.96$|$10.76 & 7.34$|$12.47 & 8.13$|$12.75 & 7.83$|$14.10 & 4.21$|$4.61 & 17.78$|$20.21 & 27.66$|$33.97 & 8.65$|$10.26 \\
{[}Acc.\tnote{1} , speed, motor RPM{]} & 8.85$|$14.25 & 8.90$|$13.44 & 9.15$|$14.81 & 10.07$|$16.32 & 4.21$|$4.99 & 9.34$|$11.22 & 17.28$|$21.35 & 11.95$|$13.92 \\
{[}Acc.\tnote{1} , motor torque, motor RPM{]} & 8.01$|$10.03 & 8.45$|$17.00 & 8.62$|$14.95 & 7.88$|$14.70 & 7.88$|$10.17 & 9.12$|$11.40 & 13.18$|$17.10 &  12.09$|$14.23\\
{[}Acc.\tnote{1} , speed, motor torque, motor RPM{]} & 7.83$|$11.17 & 6.48$|$11.40 & 6.17$|$10.72 & 7.96$|$14.55 & 9.56$|$10.91 & 7.35$|$8.73 & 7.48$|$11.98 & 12.04$|$14.18 \\ \bottomrule
\end{tabular}
}
\begin{tablenotes}
\item[1] Acceleration.
\end{tablenotes}
\end{threeparttable}
\label{EVResults}
\end{table*}

\begin{figure}[!t]
  \centering
  \includegraphics[width=0.5\columnwidth]{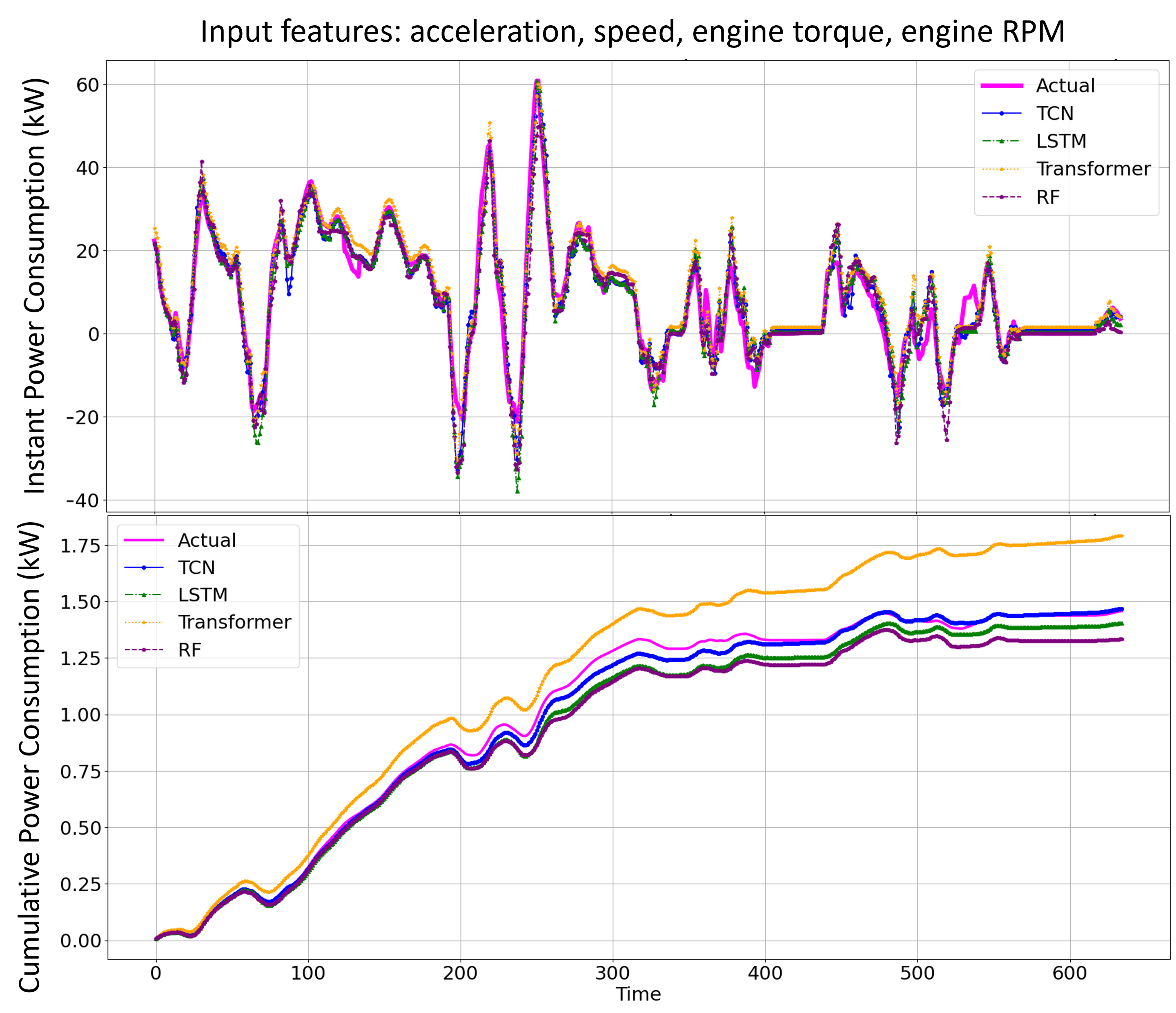}
  \caption{Instantaneous (top) and cumulative (bottom) actual power consumption and predicted values in EV dataset using TCN, LSTM, Transformer, and RF models} \label{EV_Plots}
\end{figure}

As shown in Table~\ref{EVResults}, for instantaneous power consumption with full features, [acceleration, speed, motor torque, motor RPM], Transformer achieves the best performance, yielding an MAE of 6.17 and an RMSE of 10.72. For cumulative power consumption, LSTM slightly outperforms other models with an MAE of 7.35\% and RMSE of 8.73\%. Unlike ICE vehicles, EV power consumption is influenced by motor dynamics, regenerative braking, and battery management, introducing unique temporal dependencies. Although TCN performs well for instantaneous power, it is less effective than LSTM for cumulative power estimation. RF performs well for instantaneous power prediction, particularly with the feature set \([ \text{acceleration, speed, motor torque, motor RPM} ]\), but underperforms in cumulative power estimation compared to DNNs with MAE of 12.04\% and RMSE of 14.18\% versus MAE of 7.35\% and RMSE of 8.73\% for the LSTM model. Also, Fig.~\ref{EV_Plots} presents the actual and predicted values of instantaneous and cumulative power consumption for the EV dataset, evaluated using TCN, LSTM, Transformer, and RF models.

\subsection{HEV results}
% \textcolor{blue}{For HEV I suggest to use a filter for the output so it would look smoother. The fact that we can predict the picks shows that they should be more than just noise (we cant typically learn noise) but this does not look so nice. So for the output of the HEV use another filter to smooth the result). }
\begin{table*}[!t]
\centering
\caption{Comparison of power consumption prediction using TCN, LSTM, Transformer, and RF models with different feature sets for HEV dataset.} \label{HEVResults}
\begin{threeparttable}
\resizebox{\textwidth}{!}{%
\begin{tabular}{@{}lllllllll@{}}
\toprule
 & \multicolumn{4}{c}{Instant Power Consumption Prediction} & \multicolumn{4}{c}{Cumulative Power Consumption Prediction} \\ \cmidrule(l){2-9} 
 & TCN & LSTM & Transformer & RF & TCN & LSTM & Transformer & RF \\ \midrule
 & \multicolumn{4}{c}{MAE$|$RMSE} & \multicolumn{4}{c}{MAE\%$|$RMSE\%} \\ \midrule
Acceleration & 11.27$|$15.20 & 13.19$|$18.75 & 11.71$|$15.35 & 11.74$|$15.76 & 11.92$|$13.43 & 17.59$|$18.65 & 10.81$|$14.24 & 16.77$|$19.23 \\
Speed & 8.55$|$11.83 & 8.60$|$11.97 & 13.38$|$17.22 & 12.38$|$16.56 & 19.84$|$22.06 & 9.86$|$11.82 & 9.74$|$11.99 & 7.0$|$8.02 \\
Engine RPM & 5.26$|$7.55 & 5.90$|$8.50 & 5.66$|$8.22 & 5.67$|$7.67 & 1.34$|$1.63 & 2.02$|$2.64 & 2.08$|$2.83 & 1.71$|$1.86 \\
{[}Acceleration, speed{]} & 7.98$|$11.14 & 8.52$|$11.93 & 8.58$|$11.80 & 7.23$|$10.03 & 4.57$|$5.14 & 4.84$|$5.60 & 2.72$|$3.13 & 7.41$|$8.29 \\
{[}Acceleration, engine RPM{]} & 5.56$|$7.98 & 6.14$|$8.78 & 5.28$|$7.49 & 5.07$|$6.95 & 7.40$|$8.75 & 1.73$|$2.31 & 3.55$|$3.74 & 3.00$|$3.13 \\
{[}Speed, engine RPM{]} & 5.07$|$7.04 & 5.88$|$8.64 & 5.50$|$7.60 & 5.53$|$7.76 & 1.68$|$1.79 & 1.92$|$2.19 & 11.05$|$12.14 & 2.21$|$2.53 \\
{[}Acceleration, speed, engine RPM{]} & 5.54$|$8.22 & 5.75$|$8.36 & 5.43$|$7.72 & 5.52$|$7.41 & 1.60$|$2.08 & 1.57$|$2.09 & 5.41$|$6.87 & 2.57$|$2.66 \\ 
\begin{tabular}[c]{@{}l@{}}[Acc.\tnote{1} , speed, engine torque, engine RPM]\end{tabular} & 5.59$|$7.98 & 4.72$|$6.41 & 4.95$|$7.28 & 4.88$|$6.84 & 6.04$|$6.44 & 1.13$|$1.34 & 1.57$|$2.89 & 1.55$|$1.65 \\
\bottomrule
\end{tabular}
}
\begin{tablenotes}
\item[1] Acceleration.
\end{tablenotes}
\end{threeparttable}
\end{table*}

As shown in Table~\ref{HEVResults}, less accurate predictive performance of DNNs on HEV data indicates that HEV power consumption may not fully capture the complex interactions between its mechanical and electrical components, unlike the clearer patterns observed in ICE vehicles. Since total power includes contributions from both the combustion engine and electric motor, accurate prediction relies on selecting features that capture this hybrid behavior. In terms of instantaneous power prediction, engine RPM consistently emerges as a key feature. LSTM performs best with [acceleration, speed, engine torque, engine RPM], leveraging full set of data and its capacity to model temporal dependencies with MAE of 4.72 and RMSE of 6.41. 

\begin{figure}[H]
  \centering
  \includegraphics[width=0.5\columnwidth]{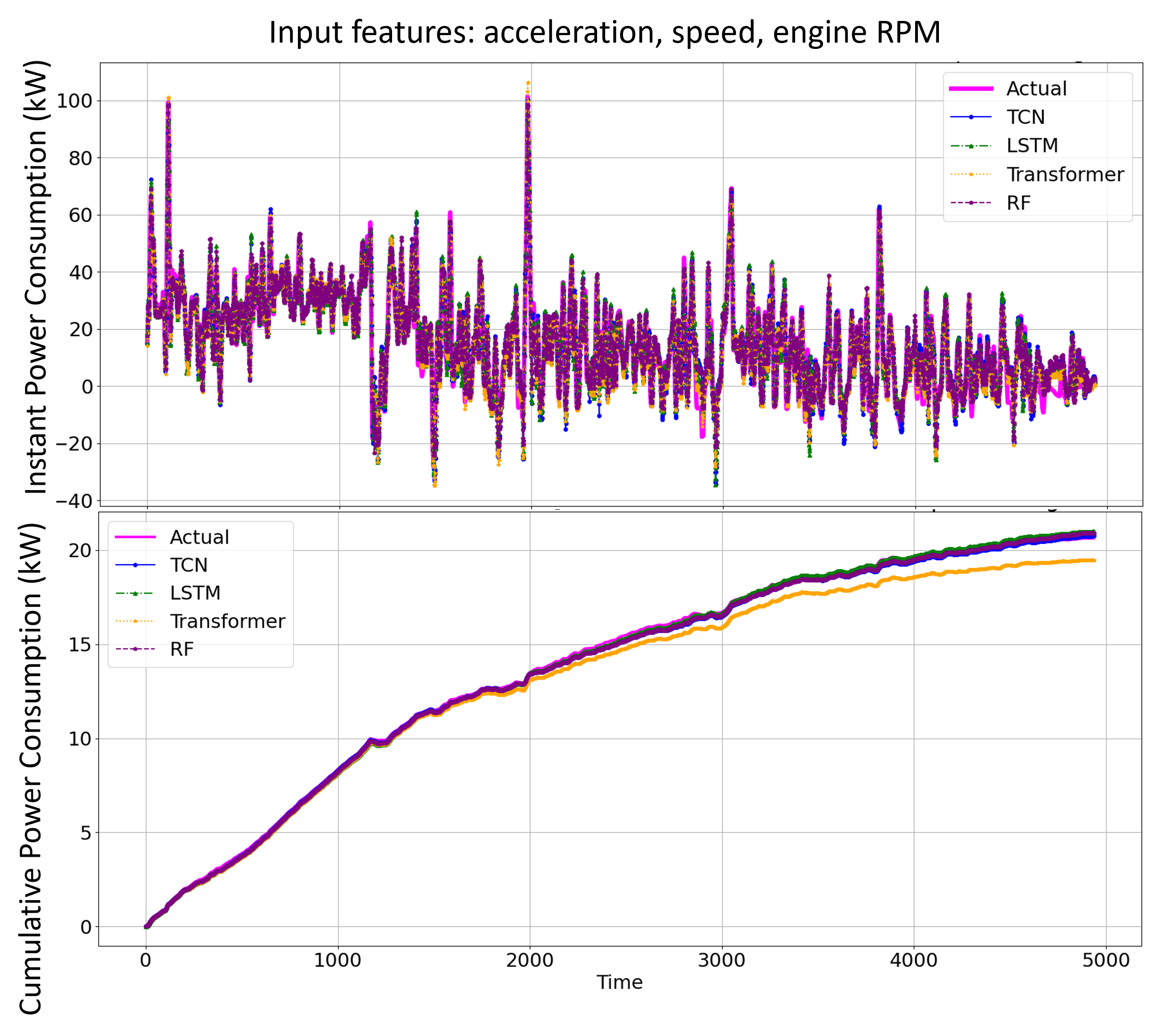}
  \caption{Instantaneous (top) and cumulative (bottom) actual power consumption and predicted values in HEV dataset using TCN, LSTM, Transformer, and RF models} \label{HEV_Plots}
\end{figure}

Regarding cumulative power prediction, both TCN and Transformer achieve strong performance when using only engine RPM. TCN yields the lowest cumulative errors with an MAE of 1.34\% and an RMSE of 1.63\%, while Transformer achieves competitive results with an MAE of 2.08\% and an RMSE of 2.83\%. However, LSTM performs best when all features are included, indicating its need for a richer context to capture long-term power trends (MAE = 1.13\%, RMSE = 1.34\%). Figure~\ref{HEV_Plots} also shows both instantaneous and cumulative power consumption, comparing actual and predicted values across the four models for the HEV dataset.
% Together, the results highlight the importance of aligning the model architecture with the prediction objectives and input features. Having said that, LSTM benefits from richer input combinations, especially in hybrid systems with multiple energy pathways.

\subsection{Computational complexity and inference time}
Table~\ref{Parameters} summarizes the computational complexity and inference time of all evaluated models. Notably, although the RF model has a relatively large number of parameters, its inference time remains the lowest due to its reliance on simple comparison operations rather than matrix computations. This highlights a key distinction between traditional ML models and DNN architectures, where model size does not always directly translate to computational cost at inference.

\begin{table}[]
\centering
\caption{Inference time for single feature entry and window size of 10.}\label{Parameters}
\begin{tabular}{@{}llll@{}}
\toprule
Model       & \#of Parameters & \#of FLOPs     & Inference Time \\ \midrule
TCN         & 104 K         &  3.2 M       &  0.47 ms             \\
LSTM        & 22 K          &  1.2 M   & 0.09 ms           \\
Transformer & 1.9 M       & 35 M   & 1.03 ms      \\ 
RF &  419 K      &  2 K  &    0.01 ms   \\ \bottomrule
\end{tabular}
\end{table}

\subsection{Uncertainty analysis}

\begin{figure}[H] 
\begin{center}
\centerline{\includegraphics[width=0.5\columnwidth]{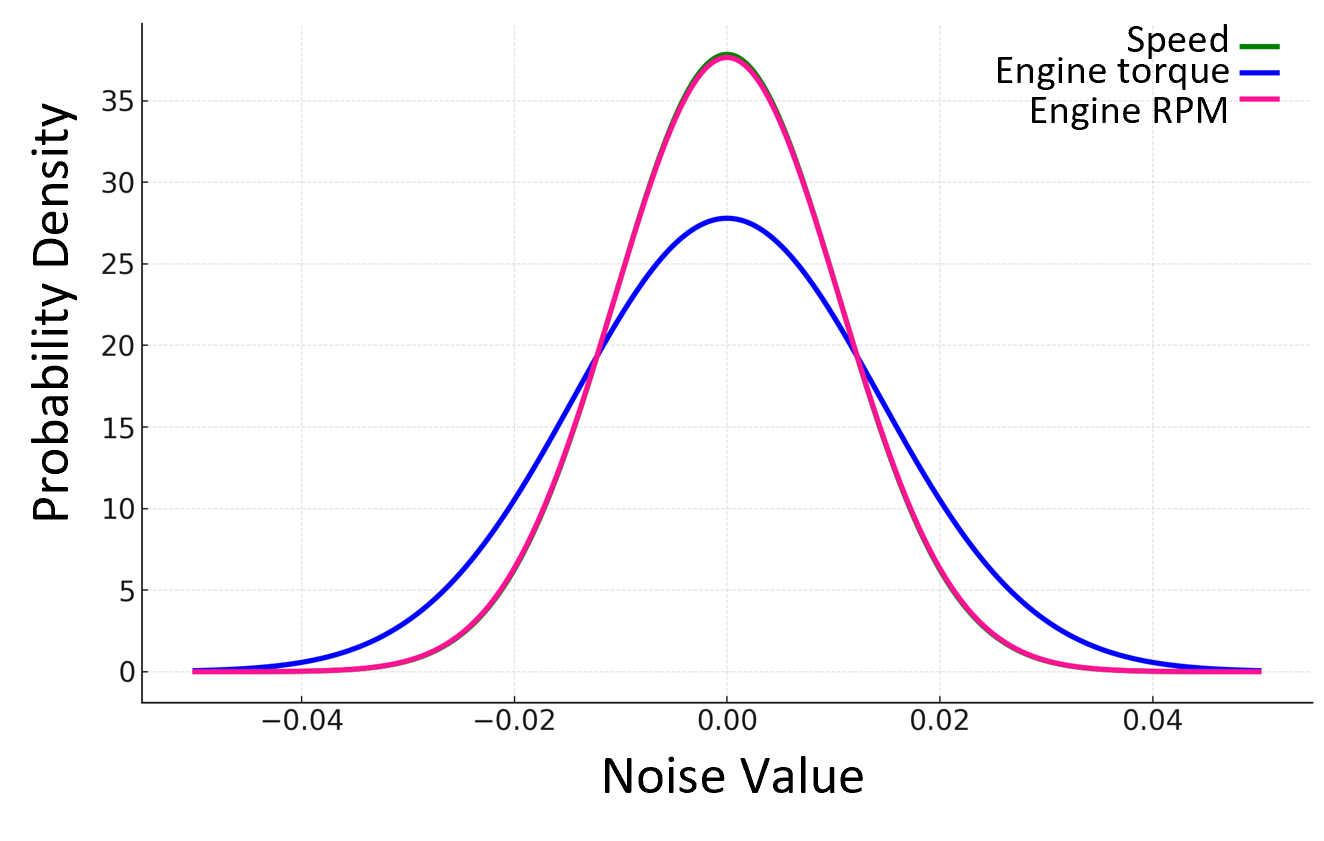}}
\caption{Feature-wise Gaussian noise modeling of ICE dataset.}
\label{Gaussian}
\end{center}
\end{figure}

\begin{figure}[!t]
\centering
\includegraphics[width=0.55\columnwidth]{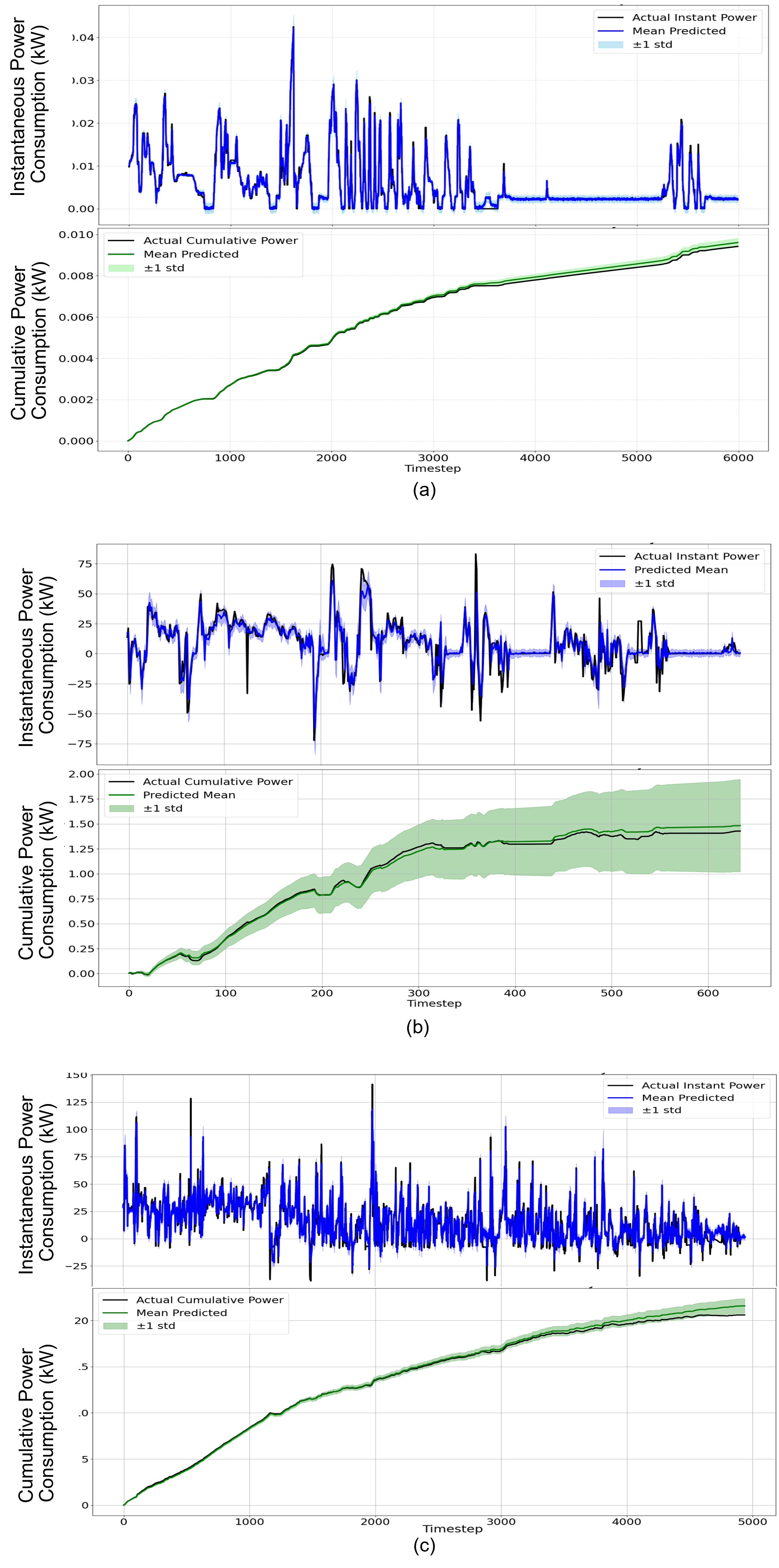}
\vspace{-0.1 in}
\caption{Uncertainty analysis: mean power consumption prediction corresponding to (a) TCN on ICE dataset, (b) LSTM on EV dataset, and (c) LSTM on HEV dataset for both instantaneous and cumulative power predictions.}
\label{UncertaintyVertical}
\end{figure}

For modeling the feature-wise Gaussian noise, we calculated the standard division of steady state segment (with length of 200 samples) of each involved feature. For example, Fig.~\ref{Gaussian} illustrates the Gaussian noise distributions of three input signals from the ICE dataset. It is evident that the distributions of speed and engine RPM are very similar. Also, dropout = 0.2 was enabled in inference time. In frequentist approach, for each experiment, we conducted \(N = 30\) runs. At each timestep, we calculated the mean and standard deviation (std) of the predictions to assess uncertainty for both instantaneous and cumulative power estimates. RF naturally quantifies uncertainty via bootstrapping across the ensemble of decision trees. As Figure \ref{UncertaintyVertical} illustrates, uncertainty bands (\(\pm 1\) std) were visualized around the mean predictions. 

Results in Table~\ref{Uncertainty} show that the EV dataset (with a standard deviation of 0.63 in MAE and 0.75 in RMSE for instantaneous power, and 5.62\% in MAE\% and 6.33\% in RMSE\% for cumulative power consumption) and the HEV dataset (with a standard deviation of 0.11 in MAE and 0.12 in RMSE for instantaneous power, and 1.27\% in MAE\% and 2.19\% in RMSE\% for cumulative power consumption) exhibited broader uncertainty margins compared to the ICE dataset (with zero standard deviation in MAE and RMSE for instantaneous power, and 0.52\% in MAE\% and 0.82\% in RMSE\% for cumulative power consumption). This aligns with the more complex power management behaviors of EVs and HEVs, such as regenerative braking and hybrid mode switching. These mechanisms introduce non-linear and state-dependent control logic, which can lead to abrupt or context-specific changes in power consumption. As a result, DNNs face greater difficulty in consistently capturing these variations, thereby leading to a wider spread of uncertainty in their predictions. 

One possible reason for the higher uncertainty observed in the EV dataset is the sensitivity of the Transformer model to input variability and noise. Transformers, due to their self-attention mechanism, can amplify small fluctuations in the input signals, especially when dropout is applied during inference. This effect is further intensified by the limited number of EV training samples (2,275) compared to the larger HEV training set (17,500), which may result in underfitting that is, the model’s inability to fully learn the underlying patterns in the data, leading to increased variability in predictions. Although RF inherently supports uncertainty estimation via bootstrapping, it exhibited higher standard deviation in predictions compared to DNNs. This may be attributed to RF's lack of temporal modeling and its sensitivity to input variability across bootstrapped subsets. In contrast, DNNs benefit from sequential learning and more controlled stochasticity through dropout, resulting in smoother and more consistent predictions with lower uncertainty margins. Another factor that may have contributed to the lower uncertainty observed in the ICE dataset is the difference in data acquisition methods (see Section \ref{Dataset}). The ICE data were collected directly from the vehicle's CAN-bus, offering high-resolution signals. In contrast, the EV and HEV datasets were obtained using a smartphone-based interface, which, while practical and scalable for real-world data collection, may involve differences in sampling rate or signal resolution. 

\begin{table*}[]
\centering
\caption{Comparison of TCN, LSTM and RF (baseline) predictions' uncertainty across different vehicles' data}\label{Uncertainty}
\begin{threeparttable}
\resizebox{\textwidth}{!}{%
\begin{tabular}{lllll}
\hline
                     &                                                                     &               & Instant Power Prediction & Cumulative Power Prediction \\ \cline{4-5} 
Vehicle              & Feature                                                             & Model         & \multicolumn{1}{c}{MAE ± $\Delta$MAE$|$RMSE ± $\Delta$RMSE}         & \multicolumn{1}{c}{MAE ± $\Delta$MAE$|$RMSE ± $\Delta$RMSE}            \\ \hline
\multirow{2}{*}{ICE} & \multirow{2}{*}{{[}Speed, engine torque, engine RPM{]}}             & TCN           & 0.0004 ± 0.0000$|$0.0009 ± 0.0000      & 1.16\% ± 0.52\%$|$1.37\% ± 0.82\%         \\
                     &                                                                     & Baseline (RF) & 0.0008 ± 0.0001$|$0.0018 ± 0.0002      & 3.28\% ± 1.67\%$|$3.81\% ± 1.93\%         \\ \hline
\multirow{2}{*}{EV}  & \multirow{2}{*}{{[}Acc.\tnote{1} , speed, motor torque, motor RPM{]}} & Transformer   & 6.19 ± 0.63$|$10.82 ± 0.75     &   7.54\% ± 5.62\%$|$12.08\% ± 6.33\%  \\
                     &             & Baseline (RF) & 7.63 ± 0.55$|$13.75 ± 1.49     & 11.54\% ± 7.72\%$|$13.77\% ± 8.33\%       \\ \hline
\multirow{2}{*}{HEV} & \multirow{2}{*}{[Acc.\tnote{1} , speed, engine torque, engine RPM]}              & LSTM          & 4.48 ± 0.11$|$6.82 ± 0.12      & 1.61\% ± 1.27\%$|$1.68\% ± 2.19\%         \\    &            & Baseline (RF) & 5.51 ± 0.11$|$7.59 ± 0.14      & 3.37\% ± 1.13\%$|$4.67\% ± 1.33\%         \\ \hline
\end{tabular}
}
\begin{tablenotes}
\item[1] Acceleration.
\end{tablenotes}
\end{threeparttable}
\end{table*}

% In terms of RF, while the result indicates the ability of RF to learn the data pattern, it generally performs worse compared to sequential models, especially for the prediction of cumulative power. For example, using the feature set \([ \text{speed, engine RPM, engine torque} ]\), RF yields an MAE of \(0.7 \times 10^{-3}\) and an RMSE of \(1.5 \times 10^{-3}\) for instantaneous prediction, and MAE and RMSE percentages for cumulative power are much higher than those of DL models. For instantaneous power prediction, DL models, particularly TCN and LSTM, offer better performance than RF, particularly when using feature sets that include engine torque and speed. For cumulative power consumption, the TCN and LSTM models consistently provide more accurate predictions compared to the RF models as they can capture the temporal aspect despite of RF.

\section{Discussion}\label{sec:discussion}

\subsection{Comparison of the vehicles}
In this section, we analyzed and compared the power consumption prediction across different vehicle types using a shared set of features. This comparison was conducted to ensure a consistent and unbiased evaluation of power consumption prediction models across ICE, EV, and HEV powertrains. By using a shared set of mutual features, the analysis eliminated feature-based variability and allowed for a direct performance comparison. It also highlighted the inherent differences in power consumption dynamics among vehicle types, offering insights into the generalizability of modeling approaches and revealing specific challenges such as regenerative braking in EVs and mode switching in HEVs.

As shown in Table~\ref{ThreeResults}, when using the same input features (acceleration, speed, engine/motor torque, and engine/motor RPM), the ICE dataset consistently produced the lowest error values for both instantaneous and cumulative power consumption across all models. For example, the best-performing model (TCN) achieved an MAE of 0.5\(\times10^{-3}\) and RMSE of 0.9\(\times10^{-3}\) for instantaneous prediction, while LSTM achieved 1.32\% MAE and 1.59\% RMSE for cumulative power. This indicated that ICE power consumption was easier to predict, likely due to its simpler and more linear powertrain behavior without complex energy blending or regenerative events.

In contrast, the EV dataset exhibited the highest error rates, regardless of the model used. The best instantaneous MAE and RMSE achieved (by Transformer) were 6.17 and 10.72, respectively, which were an order of magnitude higher than the ICE values. Similarly, cumulative power prediction on EV data resulted in the highest MAE\% and RMSE\%, with RF reaching up to 12.04\% MAE and 14.18\% RMSE. This suggested that EV power behavior was significantly more complex, possibly due to nonlinear torque control, rapid regenerative braking, and fewer consistent operational patterns.

Moreover, the HEV dataset showed intermediate performance, with prediction errors falling between those of the ICE and EV datasets. For instance, LSTM achieved 4.72 MAE and 6.41 RMSE for instantaneous power, and 1.13\% MAE and 1.34\% RMSE for cumulative prediction. While HEVs involved more complex power interactions due to their dual power sources, their cumulative signals may have been more predictable due to smoother transitions between engine and motor modes.
% \usepackage{booktabs}
\begin{table*}[!t]
\tiny
\caption{Comparison of models' performance for three vehicle types with the shared set of features ([acceleration, speed, engine/motor torque, engine/motor RPM]).}\label{ThreeResults}
\centering
\resizebox{\textwidth}{!}{%
\begin{tabular}{@{}lcccccccc@{}}
\toprule
        & \multicolumn{4}{c}{Instant Power Consumption Prediction} & \multicolumn{4}{c}{Cumulative Power Consumption Prediction} \\ \cmidrule(l){2-9} 
        & TCN            & LSTM            & Transformer     & RF        & TCN             & LSTM             & Transformer       & RF       \\ \midrule
Vehicle & \multicolumn{4}{c}{MAE$|$RMSE (For ICE: $\times10^{-3}$)}                             & \multicolumn{4}{c}{MAE\%$|$RMSE\%}                                \\ \midrule
ICE   & 0.5$|$0.9 & 0.6$|$0.9 & 1.1$|$1.3 & 0.7$|$1.5 & 2.41$|$2.88 & 1.32$|$1.59 & 1.40$|$1.67 & 3.05$|$3.58 \\
EV      &  7.83$|$11.17 & 6.48$|$11.40 & 6.17$|$10.72 & 7.96$|$14.55 & 9.56$|$10.91 & 7.35$|$08.73 & 7.48$|$11.98 & 12.04$|$14.18 \\
HEV     &  5.59$|$7.98 & 4.72$|$6.41 & 4.95$|$7.28 & 4.88$|$6.84 & 6.04$|$6.44 & 1.13$|$1.34 & 1.57$|$2.89 & 1.55$|$1.65 \\ \bottomrule
\end{tabular}
}
\end{table*}

\subsection{Comparison with other studies} \label{sec:other-studies}
In this section, we compared the proposed models with related studies for each vehicle type (ICE, EV, and HEV). For consistency and a fair comparison, we implemented the models proposed in these studies and applied them to our own datasets.
 
\subsubsection{ICE Dataset}
As shown in Table~\ref{Comparison}, when we applied FuelNet \citep{wang2023predictability}, which is a lightweight LSTM-based model for predicting vehicle power consumption from driving time series (speed and acceleration), to our ICE dataset, achieved an MAE of 0.0026 and RMSE of 0.0042 for instantaneous power prediction. However, it exhibited much higher cumulative power errors (MAE\% = 25.72, RMSE\% = 29.12). In comparison, our LSTM model, which incorporates additional engine-related features such as engine RPM and torque, substantially reduced both instantaneous (MAE = 0.0006, RMSE = 0.0009) and cumulative power prediction errors (MAE\% = 1.32, RMSE\% = 1.59), demonstrating the advantage of including engine-specific parameters in ICE energy modeling. Moreover, GBM model \citep{heni2023measuring}, which is tuned with parameters: a maximum depth of \(d = 9\), number of iterations \(I = 250\), and learning rate \(\alpha = 0.07\), using driving behavior features including speed, acceleration, travelled distance, and a stop-and-go indicator achieved lower cumulative errors than FuelNet (MAE\% = 0.65, RMSE\% = 0.78) on our dataset. While our LSTM model achieves superior performance over GBM in instantaneous power prediction, the GBM model outperforms it in cumulative power estimation. This comparison highlights the strengths and trade-offs between DNNs and classical ML models. Nonetheless, the overall strong performance of our LSTM, particularly in capturing fine-grained temporal variations, underscores the value of leveraging temporally structured features for ICE power consumption modeling.

\subsubsection{EV Dataset}
As shown in Table~\ref{Comparison}, we compared our proposed models to the The NN based model introduced by \cite{nabi2023parametric} utilizes SOC, battery and motor power, vehicle speed and distance traveled as input features. According to Table~\ref{Comparison}, compared to Transformer, this model resulted in considerably higher errors in our dataset, with an MAE of 16.84 and RMSE of 21.03 for instantaneous, and MAE\% of 79.71\% and RMSE\% of 93.23\% for cumulative power prediction. On the other hand, RF model developed by \cite{achariyaviriya2023estimating}, which estimates EV energy consumption using battery current, acceleration, speed, and road slope as input features. Their RF achieved an MAE of 8.48 and RMSE of 15.34 for instantaneous, and MAE of 6.95\% and RMSE of 7.87\% for cumulative power prediction on our dataset. However, our Transformer model, using the same input features, outperformed their RF in instantaneous power prediction (MAE = 5.19, RMSE = 9.41), though that model slightly outperformed in cumulative RMSE. These results highlight the effectiveness of our Transformer model, which achieved substantially lower errors across instantaneous prediction using only powertrain dynamic features ([acceleration, speed, motor torque, motor RPM]).

\subsubsection{HEV Dataset}
To perform a comparative analysis, we applied LSSVM (least-square SVM) \citep{zeng2018modelling} and the CNN model presented by \cite{estrada2023deep} to our dataset. Although these models rely on either pollutant estimation or intermediate variables (voltage and current) to compute energy consumption, we directly predicted total power as the sum of engine and EV contributions, using a broader set of input features. As summarized in Table~\ref{Comparison}, our LSTM model achieved the best performance among all models on the HEV dataset, with an MAE of 5.75 and RMSE of 8.36 for instantaneous power estimation, and the lowest errors (MAE\% = 1.57, RMSE\% = 2.09) for cumulative power prediction. It significantly outperforms the CNN (MAE\% = 11.65, RMSE\% = 12.71) and LSSVM (MAE\% = 6.35, RMSE\% = 5.25) models.

\begin{table*}[]
\caption{Comparison of proposed models' performance with the related studies for each vehicle type.}\label{Comparison}
\centering
\resizebox{\textwidth}{!}{%
\begin{tabular}{@{}lllllll@{}}
\toprule
\multirow{2}{*}{Dataset} & \multirow{2}{*}{Model} & \multirow{2}{*}{Input Features} & \multicolumn{2}{c}{Instant} & \multicolumn{2}{c}{Cumulative} \\ \cmidrule(l){4-7}  &       &         & MAE          & RMSE         & MAE\%         & RMSE\%    \\ \midrule
\multirow{3}{*}{ICE}     
& FuelNet \cite{wang2023predictability}       & [acceleration, speed]    & 0.0026       & 0.0042       & 25.72         & 29.12       \\ \cdashline{2-7}
& GBM \cite{heni2023measuring}               & [acceleration, speed, travelled distance, stop-and-go]             & 0.0028       & 0.0052       & 0.65          & 0.78           \\ \cdashline{2-7}   
& LSTM (Ours)  & [acceleration, speed, engine torque, engine RPM]    & 0.0006       & 0.0009       & 1.32          & 1.59  \\ \midrule

\multirow{4}{*}{EV}      
& NN \cite{nabi2023parametric}    & [SOC, travelled distance, energy consumption]                      & 16.84        & 21.03        & 79.71         & 93.23           \\ \cdashline{2-7} 
& RF \cite{achariyaviriya2023estimating}     & [battery current, acceleration, speed, road slope]                          & 8.48         & 15.34        & 6.95          & 7.87           \\ \cdashline{2-7}
& Transformer (Ours, powertrain dynamics Only)          & [acceleration, speed, motor torque, motor RPM]                              & 6.17         & 10.72        & 7.48          & 11.98           \\ \cdashline{2-7}
& Transformer (Ours)                         & [battery current, acceleration, speed, road slope]                          & 5.19         & 9.41        & 6.11          & 9.96           \\ \midrule

\multirow{3}{*}{HEV}     
& CNN \cite{estrada2023deep}                & [air mass flow, engine torque, speed, throttle position]           & 8.33         & 11.33        & 11.65         & 12.71           \\ \cdashline{2-7} 
& LSSVM \cite{zeng2018modelling}            & [motor torque, speed]                                              & 7.32         & 11.14        & 6.35          & 5.25           \\ \cdashline{2-7} 
& LSTM (Ours)                                & [acceleration, speed, engine RPM]                                  & 5.75         & 8.36         & 1.57          & 2.09           \\ \bottomrule
\end{tabular}
}
\end{table*}

\subsection{Powertrain dynamics vs. battery signals}
Data-driven methods offer several advantages: they are simpler to implement than complex physical models, adaptive to different driving patterns, capable of learning from data without requiring handcrafted rules, and, with DNNs, are well-suited for capturing temporal aspect in time series driving data. In this study, we proposed a data-driven framework to predict vehicle power consumption across ICE, HEV, and EV platforms using \textbf{only} powertrain dynamic features including acceleration, speed, engine or motor RPM, and torque. Battery-related electrical parameters such as SOC, voltage, and current are intentionally excluded from the input space. In our data-driven setting, including these parameters is not meaningful, as power can be explicitly calculated from voltage and current (\( P = V \times I \), where \(P\) is power, \(V\) is voltage, and \(I\) is current), thereby rendering the learning task trivial and negating the purpose of using a ML model. Moreover, SoC is inherently related to power consumption over time, since sustained power usage depletes battery charge, introducing a strong dependency between SoC and power that undermines the independence and generalizability of the predictive model. Furthermore, such features are not consistently available across all three vehicle types, limiting model applicability. In contrast, by relying solely on powertrain dynamic signals, the proposed approach preserves generalizability and robustness.

\section{Conclusion and Future Work}\label{sec:conclusion}
% \textcolor{blue}{Refine this section and maybe explain in a very short paragraph what can be done next? if there was any limitation you should also mention.}
In this study, we compared the performance of three DNNs (TCN, LSTM, Transformer) and a traditional ML algorithm (RF) for predicting instantaneous and cumulative power consumption across three vehicle powertrains: ICE, EV, and HEV. The results demonstrated that data-driven models can predict ICE power consumption with high precision, achieving MAE and RMSE values on the order of $10^{-3}$. This reflects the relatively stable and direct relationship between powertrain dynamic features and fuel usage in ICE vehicles. For EVs and HEVs, the prediction errors were notably higher, as reflected in both instantaneous and cumulative metrics. These higher error rates are consistent with the more complex energy management mechanisms in electrified powertrains, including regenerative braking in EVs and dynamic power distribution between engine and motor in HEVs. Nevertheless, our models, particularly LSTM and Transformer, outperformed previously proposed methods in the literature, confirming that while EV and HEV systems present greater modeling challenges, DNNs remain effective when supplied with informative features. Uncertainty analysis illustrated that EVs posing the greatest prediction challenges. The increasing uncertainty in cumulative EV power prediction over time further highlights the importance of integrating uncertainty-aware mechanisms into energy management strategies, especially for EVs and HEVs where reliability and safety depend on precise consumption forecasting.

Extending data collection to include diverse driving scenarios, such as urban and highway environments, could help assess the generalizability of the model. Incorporating longer driving sessions and a wider range of road types would further strengthen the robustness of the model. Having said that, future work could focus on moving beyond standard OBD-II interfaces by collecting high-resolution data directly from embedded vehicle sensors. This approach could provide access to more detailed signals and control-level variables, thereby improving model accuracy and enabling more advanced energy management applications.
In addition, while our study demonstrated the effectiveness of data-driven deep learning models in predicting instantaneous and cumulative vehicle power consumption, incorporating physics-informed neural networks (PINNs) or hybrid physical-data models could enhance model generalizability and interpretability by embedding physical vehicle dynamics (\textit{e.g.}, Newtonian motion, energy conservation) into the learning process.

% Importantly, moving beyond standard OBD-II interfaces and collecting high-resolution data directly from different sensors embedded in the vehicle could unlock access to more detailed signals and control-level variables, potentially enhancing model accuracy and enabling more sophisticated energy management applications. 

\section*{Acknowledgements}
The research described in this paper was jointly funded by the Clean and Energy-efficient Transportation (CEET) program of the National Research Council Canada (NRC), and the  ecoTECHNOLOGY for Vehicles ( eTV ) program of Transport Canada (TC).

\bibliographystyle{plainnat}
\bibliography{Ref.bib}

% %% For numbered reference style

%   \textit{\LaTeX: a document preparation system},
%   Addison Wesley, Massachusetts,

% \end{thebibliography}
\end{document}